\documentclass[preprint,authoryear,11pt]{elsarticle}
\usepackage[charter]{mathdesign}
\usepackage{color}
\usepackage[
  left=2.5cm,
  right=2.5cm,
  top=3cm,
  bottom=2.8cm
]{geometry}
\usepackage{amsmath}
\usepackage{amsthm}
\usepackage{soul}
\usepackage{graphicx}
\usepackage{rotating}
\usepackage{indentfirst}
\usepackage{setspace}
\usepackage{footmisc}
\usepackage{paralist}%
\usepackage{multirow} 
\usepackage{multicol}%
\usepackage{booktabs}
\usepackage{multirow}
\usepackage{threeparttable}
\usepackage{caption}
\makeatother
\allowdisplaybreaks[4]

\usepackage{subcaption}
\usepackage{bbding}
\usepackage{cite}

\usepackage[section]{placeins}
\usepackage{amsthm}
\usepackage{array}
\usepackage{natbib}
\usepackage{afterpage}
\usepackage{tabularx}
\usepackage{booktabs}
\usepackage{hyperref}
\usepackage{appendix}
\usepackage{tikz}
\usetikzlibrary{patterns}
\usetikzlibrary{calc}
\usepackage[section]{algorithm}
\usepackage{algorithmic}
\usepackage{cases}
\usepackage[mathcal]{eucal}
\usepackage{lscape}
\theoremstyle{definition}

\definecolor{rred}{RGB}{204,0,0}
\definecolor{ggreen}{RGB}{0,145,0}
\definecolor{yyellow}{RGB}{255,185,0}
\definecolor{bblue}{rgb}{0.2,0.2,0.7}

\usepackage{amsthm}

\newcounter{subsubsubsection}[subsubsection]
\renewcommand{\thesubsubsubsection}{\alph{subsubsubsection}} 

\newcommand{\subsubsubsection}[1]{%
  \refstepcounter{subsubsubsection}%
  \textit{\thesubsubsubsection)} #1%
}

\journal{}
\date{}
\makeatletter
\def\ps@pprintTitle{%
	\let\@oddhead\@empty
	\let\@evenhead\@empty
	\let\@oddfoot\@empty
	\let\@evenfoot\@empty
}
\makeatother

\begin{document}
\begin{frontmatter}

\title{Integrated Order Dispatching and Routing for Last-Mile Pickup via Deep Reinforcement Learning}
\author[a]{Yida Xu}
\ead{xuyida@tju.edu.cn}
\author[a,b]{Zhaofang Mao}
\ead{maozhaofang@tju.edu.cn}
\author[a]{Yuheng Miao}
\ead{3019209337@tju.edu.cn}
\author[a]{Jiaxin Zhang\corref{cor1}}
\ead{jiaxin.y.zhang@outlook.com}
\author[a]{Yiting Sun}
\ead{syt209054@tju.edu.cn}

\cortext[cor1]{Corresponding author. Email address: jiaxin.y.zhang@outlook.com (Jiaxin Zhang)}
\address[a]{College of Management and Economics, Tianjin University, Tianjin 300072, China}
\address[b]{Laboratory of Computation and Analytics of Complex Management Systems (CACMS), Tianjin University, Tianjin 300072, China}

\begin{spacing}{1}
\begin{abstract}
\setstretch{1.12}
	{In recent years, the growing complexity of last-mile pickup operations has increased the need for fast and accurate decision-making on logistics platforms. 
    This challenge is fundamentally driven by two key and tightly coupled decision-making processes: order dispatching and routing. 
    Solving them separately overlooks their interdependence, while fully end-to-end learning can be unstable and costly on large, variable-scale instances due to sparse rewards. 
    To solve this problem, we propose an integrated optimization framework which couples a learned routing oracle with real-time dispatching heuristics. 
    For the routing subproblem, we develop a Dynamic-Residual Graph Attention Network encoder with a Look-Ahead Courier-Personalized decoder.
    For the dispatching subproblem, we develop a routing-oracle-guided dispatching heuristic with local search, where the oracle provides near-optimal solutions to select candidate couriers while retaining real-time scalability.
    Extensive experiments on real-world datasets from Cainiao Logistics are used to test the performance of our approach, including an offline evaluation and an online rolling-horizon simulation. 
    The experimental results show that our approach outperforms other benchmarks regarding solution quality and solving time, indicating it can effectively support logistics companies in solving real-time and large-scale last-mile pickup problems.}
\end{abstract}
\end{spacing}
\end{frontmatter}


\section{Introduction}
Last-mile pickup has become a significant bottleneck within urban logistics, highlighting the importance of an effective decision-making platform as a critical focus in transportation system research \citep{ozarik2024machine, fatehi2022crowdsourcing}. 
The increasing customer demand for timely pickups pressures these platforms to make rapid decisions, typically within seconds, involving two tightly coupled processes: \textit{order dispatching} and \textit{routing} \citep{mao2025faster}. 
Order dispatching determines the assignment of accumulated orders to candidate couriers, considering constraints such as time windows, geographical proximity, and courier availability, while routing sequences the assigned pickup locations for each courier to minimize travel time and service penalties.
At each dispatching wave (e.g., every 5 minutes), the platform faces a dynamic decision scenario: couriers are distributed across the service region, each carrying previously dispatched but uncompleted orders, and the platform must dispatch newly accumulated orders to these couriers while generating feasible pickup routes.
These two processes are inherently interdependent. 
On the one hand, dispatching directly shapes each courier’s workload and spatial distribution, which in turn determine the feasibility and efficiency of subsequent routing.
On the other hand, routing outcomes, such as estimated completion times and time-window violations, provide critical feedback for evaluating dispatch quality.
Moreover, the dynamic nature of urban traffic \citep{rahmani2024toward}, heterogeneity in courier attributes \citep{pegado2024predictive}, and environmental factors such as weather conditions \citep{ozarik2024machine} introduce significant complexity and variability into this integrated decision-making process.

Due to the high complexity and computational burden of solving order dispatching and routing problem in an integrated manner, most existing studies solve these two problem separately (e.g., \citet{chen2024order}; \citet{wen2022graph2route}). While this separation improves tractability, it often leads to suboptimal decisions by failing to capture the interdependencies between dispatching and routing \citep{liu2023demand}.
Moreover, although traditional optimization methods have been widely adopted to enhance decision-making efficiency in last-mile logistics, they typically struggle to exploit historical data effectively and scale poorly in the presence of the complexity and volatility inherent in real-world environments \citep{wang2023joint}. As a result, such approaches may incur inefficiencies and higher operational costs in large-scale and dynamic settings.

In recent years, machine learning (ML) methods, particularly deep reinforcement learning (DRL), have emerged as promising approaches to address the uncertainty and dynamism of such optimization problems \citep{zhou2024learning,baty2024combinatorial}, as they can learn from historical data and adapt to dynamic environments \citep{xiao2024adaptive}. 
However, fully end-to-end DRL approaches face several challenges: they can be unstable on large, variable-scale instances due to sparse rewards \citep{mazyavkina2021reinforcement}, struggle to generalize across different graph sizes \citep{le2023limits}, and require careful handling of time-dependent travel times \citep{ccimen2024dynamic, chen2025dynamic}. 
A natural alternative is to train two separate DRL models for dispatching and routing, yet this separation not only increases computational costs but also breaks the feedback loop between these interdependent processes \citep{nazari2018reinforcement}. In conclusion, current ML approaches fail to exploit the intrinsic coupling between dispatching and routing, ultimately limiting system-level performance \citep{mao2024dl}.

\begin{figure}[htb!]
	\centering
	\includegraphics[scale=0.49]{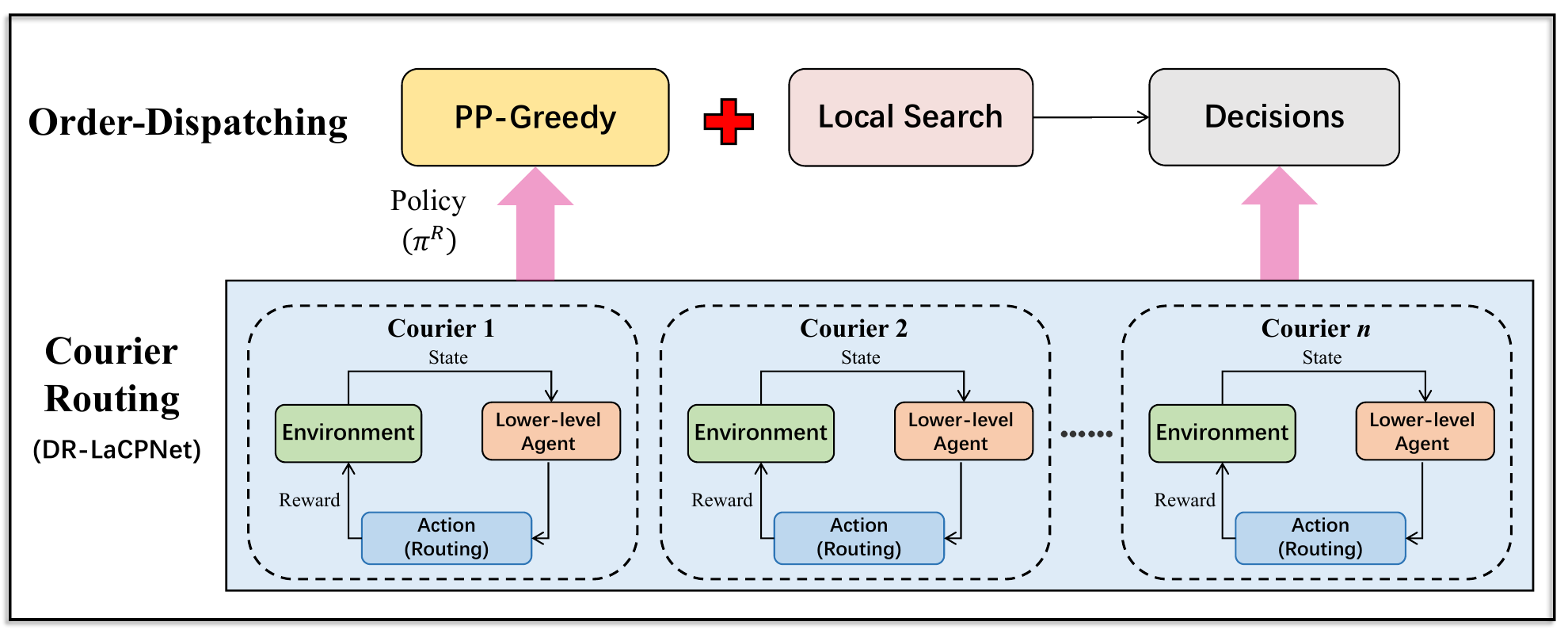}
	\caption{The proposed integrated optimization framework}
	\label{frame}
\end{figure}

To address these challenges, we develop an integrated optimization framework that couples a DRL routing oracle with lightweight dispatching heuristics.
Specifically, we first decompose the last-mile pickup problem into two hierarchical subproblems, order dispatching and routing, each formulated as a mixed-integer linear program (MILP) with time-dependent travel times.
For the routing subproblem, we recast it as a Markov decision process (MDP) and train a policy network using a rollout-baseline policy-gradient scheme. 
The policy network adopts a Dynamic-Residual Graph Attention Network encoder with a Look-Ahead Courier-Personalized decoder (DR-LaCPNet) to capture spatial-temporal dependencies and courier heterogeneity. 
The trained DRL routing oracle then guides the dispatching heuristic, which employs a position-pool screening mechanism and local search refinement to rapidly evaluate candidate couriers via routing-aware marginal cost estimation, enabling real-time decisions across variable-scale instances.
Figure \ref{frame} illustrates the decision-making process of the proposed framework.
Finally, we evaluate the proposed framework on real-world data from Cainiao Logistics (LaDe; \citet{wu2024lade}) using offline benchmarks and online rolling-horizon simulations, with sensitivity analyses on supply-to-demand ratios and objective weights.

The main contributions of this study are as follows: 
(i) To the best of our knowledge, this is the first work to address the integrated order dispatching and routing problem via DRL. We propose two novel MILP formulations with time-dependent travel times for this problem.
(ii) We develop an integrated optimization framework that couples a DRL routing oracle with dispatching heuristics, enabling rapid and accurate evaluation of dispatching through routing-aware marginal cost estimation.
(iii) For the routing subproblem, we propose the DR-LaCPNet model, where the Look-Ahead mechanism captures future time-window information and the Courier-Personalized mechanism models individual courier heterogeneity.
(iv) Extensive experiments on real-world data demonstrate that the proposed framework delivers high-quality solutions across variable-scale instances within strict time constraints, with ablation studies confirming the effectiveness of each component.

The remainder of this paper is organized as follows: 
Section \ref{liter} reviews relevant literature on non-ML and ML methods for last-mile pickup and delivery;
Section \ref{mf} presents the MILP formulations for the order dispatching and routing subproblems; 
Section \ref{method} details the MDP formulation, the DR-LaCPNet model, and the dispatching heuristic;
Section \ref{numerical} reports experimental results using offline evaluation and online rolling-horizon simulation, along with sensitivity analyses;
Section \ref{conclu} concludes and discusses future directions.

\section{Literature review} \label{liter}

\subsection{Non-ML methods for last-mile pickup and delivery} \label{oc}


\subsubsection{Order dispatching optimization}
Traditional approaches for solving the order dispatching problem often formulate it as a combinatorial optimization task, while the order pool accumulates the orders in the last dispatching wave and previous remaining orders. Among relevant studies, the dispatching process is often formulated as an MILP model or as a bipartite matching problem (BMP). For MILP models, \citet{liu2023demand} propose an MILP model for the stochastic dynamic driver dispatching and routing problem and then develop efficient exact algorithms based on Benders decomposition and column generation to solve it. For BMP models, \citet{li2022efficient} present a time-aware batch matching algorithm to offer high-quality courier-task matching in each sliding window. 

The solutions of order dispatching problems can usually be obtained through heuristic or approximate algorithms due to the computational complexity of exact solvers. The proposed heuristic rules mainly include nearest-neighbor dispatching \citep{sheridan2013dynamic,zhou2020two}, adaptive strategy \citep{liu2021time}, local search \citep{bonassa2023multi}, or meta-heuristics \citep{silva2003optimization}. Furthermore, some studies have considered multi-objective optimization. 
For example, \citet{kashansky2023intelligent} focus on the intelligent and sustainable transportation processes through the design of the multi-objective model for the logistic order dispatching system, then present an Integer Linear Programming (ILP) optimization model to solve it.
\citet{zhang2022dynamic} balance the pick-up distance, order destination regional demand, and driver service level as objectives, then propose a centralized matching algorithm to maximize all driver-order-pair values.

However, real-world last-mile logistics is inherently dynamic and uncertain. Orders arrive continuously throughout the day, and courier availability fluctuates due to break times, pickup progress, and personal preferences \citep{liu2023demand}. Moreover, external conditions such as traffic congestion and weather may further influence assignment decisions. To address such complexities, \citet{wong2006optimal} formulate a rolling horizon approach of taxi dispatching which takes the stochastic and dynamic nature of the problem into account. \citet{auad2023courier} propose an optimization-based rolling horizon algorithm for courier management that handles both region resizing and delivery task assignment decisions. Furthermore, some improvements to the conventional rolling horizon approach have been applied to dispatching problems. \citet{huang2023receding} propose a receding-horizon dispatching model that periodically re-optimizes assignments as new data becomes available. Nevertheless, these methods still largely rely on static optimization formulations, which limit their responsiveness and scalability in high-frequency dispatching scenarios.

\subsubsection{Courier routing optimization}
At each decision stage, the courier routing problem without additional constraints can be regarded as a variant of the traveling salesman problem (TSP), namely open TSP (OTSP, \citet{bellmore1974transformation}). The only difference between TSP and OTSP is that the courier does not necessarily need to return to the warehouse after picking up goods from the customer's position in the OTSP. 
When considering the time-dependent travel time, the traditional TSP would become the time-dependent traveling salesman problem (TDTSP).
Therefore, in this section, we review the relevant literature on non-ML methods on TSP, OTSP and TDTSP.

TSP has been extensively studied in both static and dynamic settings. In the static TSP, all customer positions and inter-city distances are known and fixed prior to route planning. This classical version has been the focus of numerous algorithmic developments over the past decades \citep{voudouris1999guided, englot2013efficient, aksit2024flying}. 
In contrast, the dynamic TSP (DTSP) introduces real-time elements such as changing travel times, newly appearing positions, or evolving cost structures, which require adaptive planning throughout the tour \citep{chang2009stochastic, toriello2014dynamic, mavrovouniotis2016ant, unal2023dynamic}. 
Several non-ML approaches have been developed to solve this NP-hard problem. For heuristics or meta-heuristics, \citet{theys2010using} use Lin–Kernighan–Helsgaun heuristic to solve the TSP and verify whether the sophistication of ``state-of-the-art'' local search heuristics is necessary for routing order pickers in warehouses. Exact algorithms with tailored cuts also show great efficiency when dealing with TSP variants. For example, \citet{cha2024exact} introduce a TSP with a sequence-and-load dependent risk and propose an iterative-deepening A*-based tree search algorithm to solve it. For a more comprehensive survey of TSP variants, we refer the readers to the work of \citet{pop2024comprehensive}.

The OTSP is a fundamental variant of the classical TSP, in which the salesman is not required to return to the starting point. While the problem remains NP-hard, similar to the classical TSP, its asymmetric structure and open-ended nature require specialized solution methods \citep{frieze1982worst}. Early approaches include adaptations of branch-and-bound and dynamic programming techniques (e.g., \citet{pekny1990parallel}). Additionally, several work has focused on heuristic and approximation algorithms to solve OTSP, often tailored to handle large-scale instances or real-world constraints (e.g., \citet{glover2001construction, barketau2016approximation}).

Another important variant of TSP is to consider the time-dependent travel time, which indicates that the travel time between two nodes would vary according to the time intervals or the sequence of edges, as introduced in \citet{wang2019time}. Among related works, \citet{cordeau2014analysis} formulate the TDTSP as an integer linear programming (ILP) model for which valid inequalities are devised. \citet{pralet2023iterated} proposes an Iterated Maximum Large Neighborhood Search (ImaxLNS) to solve the TDTSP with time window (TDTSPTW). \citet{ccimen2024dynamic} develop an MILP for the open time-dependent traveling salesman problem to determine routing and vehicle speed decisions. These formulation structures are well-suited for solving using decomposition methods, but they need to be appropriately modified to be compatible with the DRL framework.

\subsection{ML methods for last-mile pickup and delivery} \label{literml}
Recent years have witnessed a growing interest in applying ML techniques to last-mile pickup and delivery. The shift is driven by the increasing complexity of urban delivery networks and the availability of granular operational data. This section reviews the recent literature on ML-based approaches for last-mile pickup and delivery problems.

A central focus in the literature is the use of ML for route prediction and dispatching optimization. \citet{mao2023drl4route} present the first attempt to generalize Reinforcement Learning (RL) to the route prediction task, leading to a novel RL-based framework called DRL4Route. Also, \citet{mo2023predicting} propose a pair-wise attention-based pointer neural network to model the actual stop sequences followed by couriers, leveraging historical GPS traces. Their model significantly outperforms conventional shortest-path heuristics, highlighting the limitations of purely distance-based metrics in real-world operations. Building on this direction, \citet{denis2025papn} introduce Proximity Attention and Pointer Network framework (PAPN), which models the decision process from courier behavior, and performs competitively with state-of-the-art reinforcement learning models. \citet{zhou2021multi} choose order fee and the spatio-temporal relationship as two objectives, then combine Implicit Quantile Networks (IQN) with the traditional Deep Q-Networks (DQN) to jointly learn the two return distributions and adjusting their weights. These studies demonstrate that incorporating domain-specific features and attention mechanisms into supervised learning frameworks improves route prediction accuracy and scalability.

Another strand of research explores hybrid ML-optimization frameworks. \citet{carvalhosa2024enhancing} develop a two-stage model that first uses a neural network to predict delivery zones and then applies a LKH-based metaheuristic for route refinement. Their approach shows that ML can act as a pre-processing tool to reduce problem complexity and enhance solver performance. Similar ideas are adopted by \citet{bruni2023machine}, who present a new heuristic algorithm to solve the variable cost and size bin packing problem with stochastic items based on machine learning techniques. However, a common limitation in these hybrid approaches is the reliance on high-quality labeled data, which may not be available in all deployment contexts.

Recent work has explored the use of graph neural networks (GNNs) to better model the structural and decision-making complexities of last-mile pickup and delivery. GNNs effectively capture spatial and relational dependencies in road networks and pickup and delivery tasks. For example, \citet{wen2022graph2route} propose a dynamic spatial-temporal graph-based model, named Graph2Route, to solve the pick-up and delivery route prediction task. \citet{chen2024order} propose an optimization algorithm based on GNN to solve the order dispatching problem. 
Despite these advances, most studies still treat dispatching and routing as separate components, leaving their strong interdependence insufficiently exploited. While integrated solutions have been attempted (e.g., hierarchical deep reinforcement learning; \citet{li2025optimizing}), fully end-to-end learning frameworks often suffer from sparse and delayed rewards. Their training and tuning costs also scale poorly with the highly variable instance size in last-mile pickup, undermining convergence stability and practical deployability.

Our motivation for adopting a DRL optimization approach stems from the limitations of existing methods for last-mile pickup problems. 
In this work, we address the integrated order dispatching and routing problem with time-dependent travel times, a real-world variant that incorporates complex spatial-temporal dynamics and stochastic operational conditions. Solving this problem efficiently at scale requires fine-grained coordination between dispatching decisions and routing outcomes. Instead of learning dispatching and routing jointly from scratch, we adopt a learning-augmented perspective by coupling a learned DRL routing oracle with lightweight dispatching heuristics. This framework enables fast, scalable, and controllable real-time operations, effectively handling the interdependence between dispatching and routing without the need for costly end-to-end training. The studies reviewed in this section are summarized and compared in Table~\ref{litertable}, providing a comprehensive overview of their characteristics and approaches.

\begin{table}[htbp]
\setstretch{1.25}
\footnotesize
\caption{An overview of recent related studies for order dispatching and courier routing}
\label{litertable}
\centering
\begin{tabular}{lcccc>{\raggedright\arraybackslash}p{3cm}}
\hline
\textbf{Study} & \textbf{Strategy} & \textbf{Order dispatching} & \textbf{Courier routing} & \textbf{TD-TT} & \textbf{Method} \\
\hline
\citet{wang2018deep} & Static & \Checkmark &  &  & DRL \\
\citet{zhou2020two} & Static & \Checkmark &  &  & Heuristics \\
\citet{liu2021time} & Dynamic & \Checkmark &  & & DL + Heuristics \\
\citet{zhou2021multi} & Static & \Checkmark & &  & Distributional RL \\
\citet{wen2022graph2route} & Dynamic &  &  &  & DL \\
\citet{li2022efficient} & Dynamic & \Checkmark & &  & BMP \\
\citet{zhang2022dynamic} & Dynamic & \Checkmark & &  & Reward Learning\\
\citet{auad2023courier} & Dynamic & \Checkmark &  &  & BMP \\
\citet{liu2023demand} & Dynamic & \Checkmark & \Checkmark & \Checkmark & Exact algorithms\\
\citet{mao2023drl4route} & Dynamic & & \Checkmark &  & DL \\
\raisebox{-.7\height}{\citet{zhen2023heterogeneous}} &
\raisebox{-.7\height}{Static} &
\raisebox{-.7\height}{\Checkmark} &
\raisebox{-.7\height}{\Checkmark} &
\raisebox{-.7\height}{\Checkmark} &
Exact + Heuristic algorithms\\
\citet{chen2024order} & Dynamic & \Checkmark &  &  & DL + Heuristics \\
\citet{ozarik2024machine} & Static & \Checkmark & &  & DL + Heuristics \\
\citet{pegado2024predictive} & Dynamic &  & \Checkmark & & DL + Heuristics \\
\citet{carvalhosa2024enhancing}  & Dynamic & & \Checkmark &  & DL + Heuristics \\
\citet{li2025optimizing} & Dynamic & \Checkmark & \Checkmark & & HDRL \\
\hline
This study & Dynamic & \Checkmark & \Checkmark & \Checkmark & DRL + Heuristics \\
\hline
\end{tabular}
\end{table}

\section{Mathematical formulation} \label{mf}
\begin{figure}[htb!]
	\centering
	\includegraphics[scale=0.48]{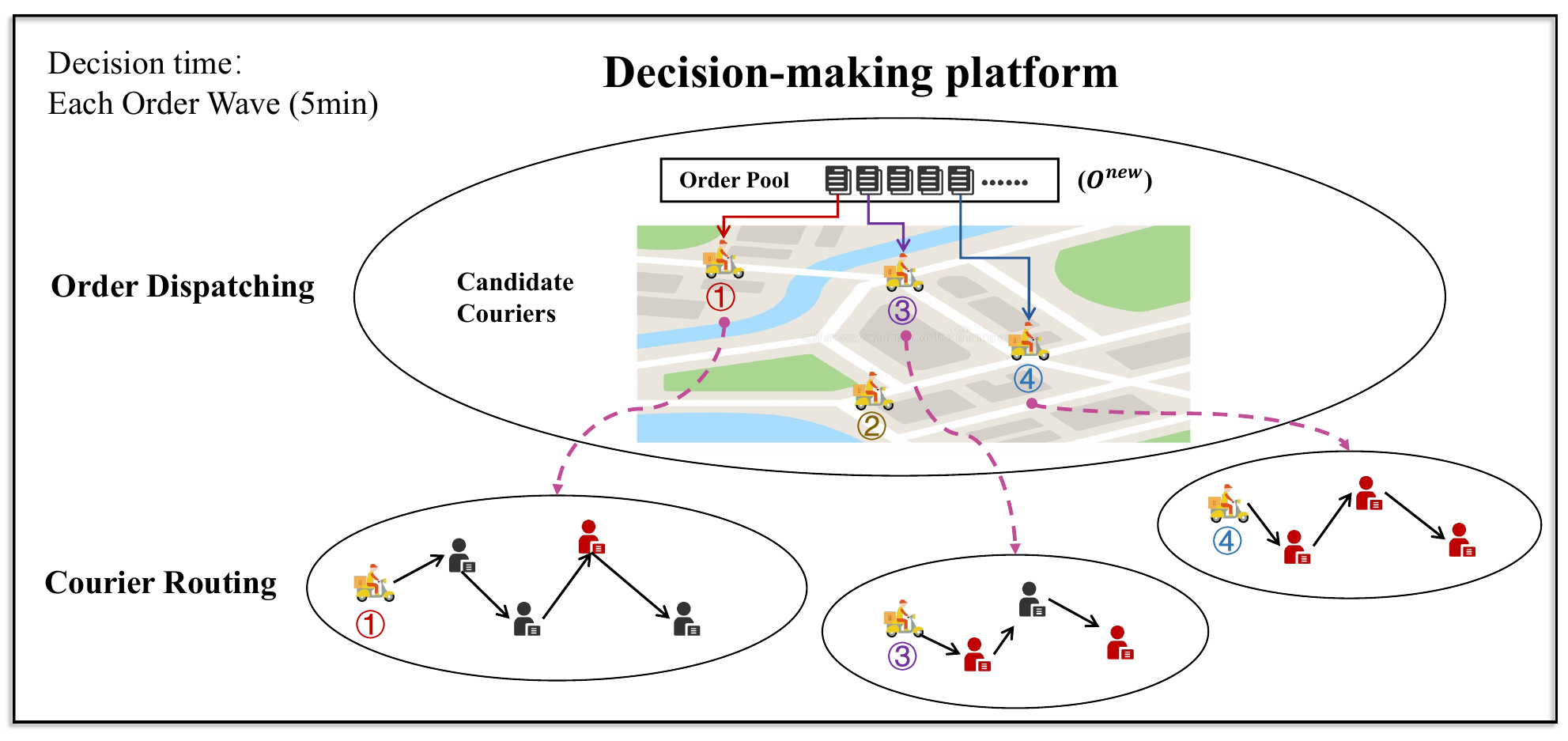}
	\caption{Schematic diagram of integrated order dispatching and routing}
	\label{description}
\end{figure}

In this paper, we address the integrated order dispatching and routing problem, from the perspective of decision-making platform of a logistics company (e.g., Cainiao Logistics). During each decision cycle, the decision-making platform would make operational decisions based on current order and courier availability, while simultaneously dispatching orders and generating reference routes for couriers. The schematic diagram of the integrated order dispatching and routing problem is illustrated in Figure \ref{description}.

To handle this challenging problem, we decompose the original problem into two distinct subproblems, the order dispatching problem and the routing problem. We formulate each of the subproblem as an MILP model. Table \ref{notations} shows the notations used for both subproblems. In the subsequent sections, we would introduce these two subproblems in detail.

\begin{table}[htbp]
\caption{The summary of notations}
\label{notations}
\begin{tabular}{lp{14cm}}
\hline
Notation & \multicolumn{1}{c}{Definition} \\ \hline
\multicolumn{2}{l}{Sets:} \\
$\mathcal{K}$ & the set of all candidate couriers \\
$\mathcal{L}$ & the set of  starting positions of all candidate couriers \\
$O^{new}$ & the set of all orders available for dispatch \\
$O^{old}$ & the set of all in-hand orders\\
\multicolumn{2}{l}{Parameters:} \\
$t_{r i j}$ & travel time from node $i$ to $j$ during time interval $r$ \\
$g_{i r}$ & the time of node $i$ at time interval $r$ when the travel time function changes \\
$a_{i},b_{i}$ & the release time and the deadline of node $i$\\
$\alpha,\phi$ & objective weights \\
\multicolumn{2}{l}{Decision and auxiliary variables:} \\
$y_{i}^{k}$ & 1, if order $i$ is dispatched to courier $k$; 0, otherwise \\
$x_{i j}^{k}$ & 1, if courier $k \in \mathcal{K}$ traverses arc $(i,j)$; 0, otherwise \\
$c_{i j}^k$ & the realized travel time from node $i$ to node $j$ of courier $k$ \\
$U_{i r}^{k}$ & 1, if courier $k$ arrive at node $i$ in time interval $r$; 0, otherwise \\
$\beta_{i}^{k}$ & the time when courier $k$ arrive at node $i$ \\
$\overline{\beta}_{i r}^{k}$ & auxiliary variable which takes value $\beta_{i}^{k}$ when $U_{i r}^{k}=1$; 0, otherwise \\ 
$p_{i}^{k}$ & Penalty at node $i$ for courier {k}\\
\hline
\end{tabular}
\end{table}

\subsection{Order dispatching subproblem} \label{ordersub}
In order to handle the dynamic order dispatching subproblem, we consider the decision-making process at each dispatching wave. We discretize the operating horizon into a series of time intervals, denoted as $\mathcal{T}=\{0, 1, 2, \cdots, T\}$, each time interval has length $\Delta t$. At each decision time, let $O^{new}$ denote the set of all orders available for dispatch, consisting of all previously undispatched orders and newly arriving orders. Let $\mathcal{K}$ be the set of all candidate couriers, each candidate courier $k \in \mathcal{K}$ carries $q_k$ in-hand orders that have been dispatched but not completed for pickup, denoted as $O^{old}_k$. Let $O^{\mathrm{old}}=\bigcup_{k\in\mathcal{K}} O_k^{\mathrm{old}}$ be the aggregated set of all such in-hand orders. Each order $i$ associated with a pickup time window $[a_i,b_i]$, any early or late pickup results in a penalty $p$. Each candidate courier $k$ is located at $l_k \in \mathcal{L}$ at the beginning of the dispatching wave, where $\mathcal{L}$ is the set of starting positions of all candidate couriers. We describe the order dispatching subproblem using $\mathcal{G}=(\mathcal{V},\mathcal{A})$, where $\mathcal{V}=\mathcal{L} \cup O^{old} \cup O^{new}$ is the set of nodes and $\mathcal{A}$ is the arc set. Note that $\mathcal{G}$ is not a complete graph, since orders that have been dispatched to courier $k$ will not be picked up by other couriers.

\begin{equation}
\mbox{minimize}\quad  \sum_{k \in \mathcal{K}}\left(\alpha \times  \mathop{\max}_{i \in \mathcal{V} \setminus \mathcal{L}}\beta_i^{k} + \phi \times \sum_{i \in \mathcal{V}\setminus \mathcal{L}} p_i^{k} \right)
\label{c0}
\end{equation}
\begin{equation}
\sum_{j \in \mathcal{V}} x_{i, j}^k=y_i^k, \quad \forall i \in O^{new}, k \in \mathcal{K}
\label{c1}
\end{equation}
\begin{equation}
\sum_{k \in \mathcal{K}} y_{i}^k=1, \quad \forall i \in O^{new}
\label{c2}
\end{equation}
\begin{equation}
x_{ij}^{k} \in \{0, 1\} \quad \forall i, j \in \mathcal{V}^{k},i \neq j
\label{c3}
\end{equation}
\begin{equation}
y_i^{k} \in \{0,1\} \quad \forall i \in \mathcal{V} 
\label{c4}
\end{equation}
where $\beta_i^{k}$ and $p_i^{k}$ are computed from the reference route of courier $k$ returned by the routing subproblem defined in Section~\ref{milp_routing}.

The objective \eqref{c0} consists of two components: (1) Minimize the total travel time; (2) Minimize the total time window penalty of all candidate couriers. Here we use adjustable weight factors $\alpha$ and $\phi$ to balance the trade-off between these two components. Constraints \eqref{c1} ensure the dispatched orders must be served. Constraints \eqref{c2} guarantee that all new orders must be dispatched to a specific courier. Finally, constraints \eqref{c3}-\eqref{c4} present the domains of $x_{ij}^{k}$ and $y_i^{k}$.

\subsection{Routing subproblem} \label{milp_routing}
Subsequently, the routing subproblem can be considered as a parallel optimization problem for each candidate courier. For simplicity, we list the routing subproblems for courier $k$ here, which we denote as time dependent traveling salesman problem with time window for pickup (TDTSPTW-PU). We formulate the TDTSPTW-PU as an MILP defined on a complete graph $\mathcal{G}^{k}=(\mathcal{V}^{k}, \mathcal{A}^{k})$ with finite node set $\mathcal{V}^{k}$ and finite arc set $\mathcal{A}^{k}$ for courier $k$. Let $l_k$ denote the starting position of courier $k$, then $\mathcal{V}^{k}= {l_k} \cup O^{old}_{k} \cup O^{new}_{k}$ and graph $\mathcal{G}^{k}$ is a subgraph of graph $\mathcal{G}$. Formally, TDTSPTW-PU can be regarded as a special case of TDTSPTW, with minor revisions to set the travel time of all nodes back to the courier's starting position (warehouse in TDTSPTW) to 0. 

Here we consider time-dependent travel time based on time intervals, following the study of \citep{wang2019time}. The time-dependent travel time matrix is denoted as $\mathbf{T} \in \mathbb{R}^{T \times \lvert \mathcal{V}^{k} \rvert \times \lvert \mathcal{V}^{k} \rvert}$ and each element of it can be represented by $t_{r i j}$, indicating the travel time from node $i$ to node $j$ at time period $r$. In order to better depict discrete travel times, we take \citet{chen2025dynamic}'s suggestion and follow \citet{kok2011optimizing}'s formulation by introducing the variables $U_{i r}^{k}$ and $\beta_i^{k}$; parameter $g_{i r}$; auxiliary variable $\overline\beta_{i,r}^{k}$ to aid in modeling. The binary variables $U_{i, r}$ takes 1 only if $g_{i, r} \leqslant \beta_i^k \leqslant g_{i, r+1}$, while $\overline\beta_{i r}^{k}$ takes the value of $\beta_{i}^k$ if the corresponding variable $U_{i r}$ is one, and zero otherwise. In this way, we avoid using the unknown time period $r$ directly as a subscript for the travel time $t_{r i j}$, transforming it into an optimization problem that can be solved through solvers. The mathematical formulation of TDTSPTW-PU is given below.

\begin{equation}
\mbox{minimize}\quad  
\left(
\alpha \times  \mathop{\max}_{i \in \mathcal{V}^{k}\setminus\{l_k\}}\beta_i^{k} 
+ 
\phi \times \sum_{i \in \mathcal{V}^{k}\setminus\{l_k\}} p_i^{k} 
\right)
\label{c10}
\end{equation}
\begin{equation}
\mbox{s.t.}\quad \sum_{i \in \mathcal{V}^{k}, i \neq j} x_{ij}^{k} = 1 \quad \forall j \in \mathcal{V}^{k} 
\label{c11}
\end{equation}
\begin{equation}
\sum_{j \in \mathcal{V}^{k}, j \neq i} x_{ij}^{k} = 1 \quad \forall i \in \mathcal{V}^{k} 
\label{c12}
\end{equation}
\begin{equation}
\sum_{i \in S} \sum_{j \in S, j \neq i} x_{ij}^{k} \leq |S| - 1 \quad \forall S \subset \mathcal{V}^{k}\setminus\{l_k\},\ 2 \le |S| \le |\mathcal{V}^{k}|-1 .
\label{c12a}
\end{equation}
\begin{equation}
\beta_j^{k} \geq \beta_i^{k} + c_{ij}^k - M(1 - x_{ij}^{k}) \quad \forall i, j \in \mathcal{V}^{k} , i \neq j, j \neq l_k \label{c13}
\end{equation}
\begin{equation}
\beta_{l_k}^{k} = 0 .
\label{c13a}
\end{equation}
\begin{equation}
\sum_{r=1}^{T} U_{i,r}^{k} = 1 \quad \forall i \in \mathcal{V}^{k} 
\label{c14}
\end{equation}
\begin{equation}
g_{i,r} U_{i,r}^{k} \leq \overline\beta_{i,r}^{k} \quad \forall i \in \mathcal{V}^{k} , r \in \mathcal{T} 
\label{c15}
\end{equation}
\begin{equation}
g_{i,r+1} U_{i,r}^{k} \geq \overline\beta_{i,r}^{k} \quad \quad \forall i \in \mathcal{V}^{k}, r \in \mathcal{T} 
\label{c16}
\end{equation}
\begin{equation}
\sum_{r=1}^{T} \overline\beta_{i,r}^{k} = \beta_i^{k} \quad \forall i \in \mathcal{V}^{k}
\label{c17}
\end{equation}
\begin{equation}
c_{i j}^{k} \geq t_{r i j} + M(U_{i r}^{k} +x_{i j}^{k}- 2) \quad \forall i,j \in \mathcal{V}^{k}, i \neq j, r \in \mathcal{T}
\label{c18}
\end{equation}
\begin{equation}
p_i^{k}=\left\{\begin{array}{cc}
\beta_i^{k}-b_i, & \beta_i^{k}  > b_i \\
a_i-\beta_i^{k}, & \beta_i^{k} < a_i \\
0, & a_i \leq \beta_i^{k} \leq b_i
\end{array}\right.
\quad \forall i \in \mathcal{V}^{k} 
\label{c19}
\end{equation}
\begin{equation}
x_{ij}^{k} \in \{0, 1\} \quad \forall i, j \in \mathcal{V}^{k},i \neq j
\label{c20}
\end{equation}
\begin{equation}
U_{ir}^{k} \in \{0, 1\} \quad \forall i \in \mathcal{V}^{k},t \in \mathcal{T}
\label{c21}
\end{equation}
\begin{equation}
c_{ij}^{k} \in \mathbb{R}_{>0} \quad \forall i, j \in \mathcal{V}^{k},i \neq j
\label{c22}
\end{equation}
\begin{equation}
\beta_i^{k} \in \mathbb{R}_{>0} \quad \forall i \in \mathcal{V}^{k} 
\label{c23}
\end{equation}
\begin{equation}
\overline\beta_{i,r}^{k} \in \mathbb{R}_{>0} \quad \forall i \in \mathcal{V}^{k} , r \in \mathcal{T}
\label{c24}
\end{equation}

As previously introduced, the objectives are to minimize the maximum arrival time and the time window penalty, as shown in \eqref{c10}. Constraints \eqref{c11}-\eqref{c12} are known as flow conservation constraints. To eliminate subtours and enforce a single tour visiting all nodes, we further introduce the Dantzig--Fulkerson--Johnson (DFJ) subtour-elimination constraints \eqref{c12a}. Constraints \eqref{c13} indicate the relationship between arrival time and travel time and constraint \eqref{c13a}. Constraints \eqref{c14}-\eqref{c17} forces $g_{i, r} \leqslant \beta_i^k \leqslant g_{i, r+1}$ if and only if $U_{i r}^{k}=1$. Constraints \eqref{c18} guarantee $c_{i j}^{k} \geq t_{r i j}$ if and only if $U_{i r}^{k}=x_{i j}^{k}=1$. Constraints \eqref{c19} characterize the nonlinear relationship between penalty $p_{i}^{k}$ and arrival time $\beta_{i}^{k}$, which can be easily generalized to other forms of time window penalty functions. Constraints \eqref{c20}-\eqref{c24} present the domain of the variables. The proposed MILPs are then served as baselines in our experiments.

\section{Methodology} \label{method}
In this section, we describe the solution methods for the routing subproblem and dispatching subproblem, respectively. 
In Section \ref{lower_model}, we first establish an MDP model and introduce the Dynamic-Residual Graph Attention Network encoder with a Look-Ahead Courier-Personalized decoder (DR-LaCPNet) for the routing subproblem. 
Consequently, in Section \ref{heuristic_upper}, we propose a position-pool greedy dispatch with local search refinement (PP-Greedy-LS) algorithm using the trained DR-LaCPNet as the DRL routing oracle.

\subsection{DRL for the routing subproblem} \label{lower_model}
In this section, we first model the routing subproblem as an MDP, then employ the reinforcement learning algorithm for training while dynamically updating the baseline. Finally, the DR-LaCPNet is proposed to capture the spatial-temporal characteristics and to make decisions sequentially.

\subsubsection{Markov Decision Process}
As stated earlier, the oprating horizon is discretized as $\mathcal{T}=\{0, 1, 2, \cdots, T\}$. Here, we model the routing subproblem as a sequential decision MDP, the decisions are made at each time step. Following the notations of \citet{chen2025dynamic}, we employ $d$ to represent the index of the current decision step. 
The MDP of the routing subproblem is described as follows.

\textbf{State}: To better demonstrate the state compositions, we divide the states $S_{d}^k$ into three distinct components for decision step $d$: static courier global information $\mathbb{G}^{k}$, environmental information $\mathbb{E}_{d}^{k}$ and dynamic courier local information $\mathbb{L}_{d}^{k}$, i.e., $S_{d}^k=(\mathbb{G}^{k}, \mathbb{E}_{d}^{k}, \mathbb{L}_{d}^{k})$. The static courier global courier information $\mathbb{G}^{k}$ consists of inherent courier attributes, including the starting position $l_k=(\overline{x}_k^{ini},\overline{y}_k^{ini})$ of courier $k$, average traveling speed $v_{k}^{avg}$, average task completion time $CT_{k}^{avg}$, courier index $NU_{k}$, and the order type list $OT_{k}$ indicating the old orders or new dispatched orders and order time windows $[a_{i}, b_{i}]$. The environmental information $\mathbb{E}^{k}_{d}$ includes timestamp $ts_{d}$, date stamp $ds_{d}$, traffic congestion factor $\xi_d$, and weather $W_{d}$. Finally, the dynamic courier local information $\mathbb{L}_{d}^{k}$ consists of current courier position $(\overline{x}_k^{d},\overline{y}_k^{d})$, the vector of visited mask $M_{k}^{d}=(v_{k, 0}^{d},v_{k,1}^{d}, \cdots, v_{k,|\mathcal{V}^{k}|}^{d})$, the vector of arrival times $AT_{k}^{d}=(\vartheta_{k,0}^{d}, \vartheta_{k,1}^{d}, \cdots, \vartheta_{k,|\mathcal{V}^{k}|}^{d})$ and the vector of rewards $\mathcal{R}_{k}^{d}=(R_{k,0}^{d}, R_{k,1}^{d}, \cdots, R_{k,|\mathcal{V}^{k}|}^{d})$.

\textbf{Action}: The action is to determine the next order to pickup, which would incur a reward according to the travel time and time window penalty.

\textbf{State Transition Rule}: The state transition rule will let the current state $S_{d}^k$ move to the next state $S_{d+1}^k$ based on the previous action. Suppose node $j$ is chosen as the next node at the decision step $d+1$ and the last visited node is $i$, then $v_{k,j}^{d+1}:=1$, $\vartheta_{k,j}^{d+1}:=\vartheta_{k,i}^{d}+t_{r_{d}^{k} i j}$, $(\overline{x}_k^{d+1}, \overline{y}_k^{d+1}):=(\overline{x}_j,\overline{y}_j)$ and $R_{k,j}^{d}:= \alpha \times t_{r_{d}^{k} i j} + \phi \times p_j $, where $r_{d}^{k}=\left\lfloor\frac{\vartheta_{k,i}^{d}}{\Delta t}\right\rfloor$ is the last time interval and $(\overline{x}_j,\overline{y}_j)$ is the coordination of order $j$.

\textbf{Objective Function}: The objective function is set to be the same as Equation \eqref{c10}, while both considering the total travel time and the time window penalty.

\subsubsection{Reinforcement learning algorithm}

\begin{figure}[htb!]
	\centering
	\includegraphics[scale=0.48]{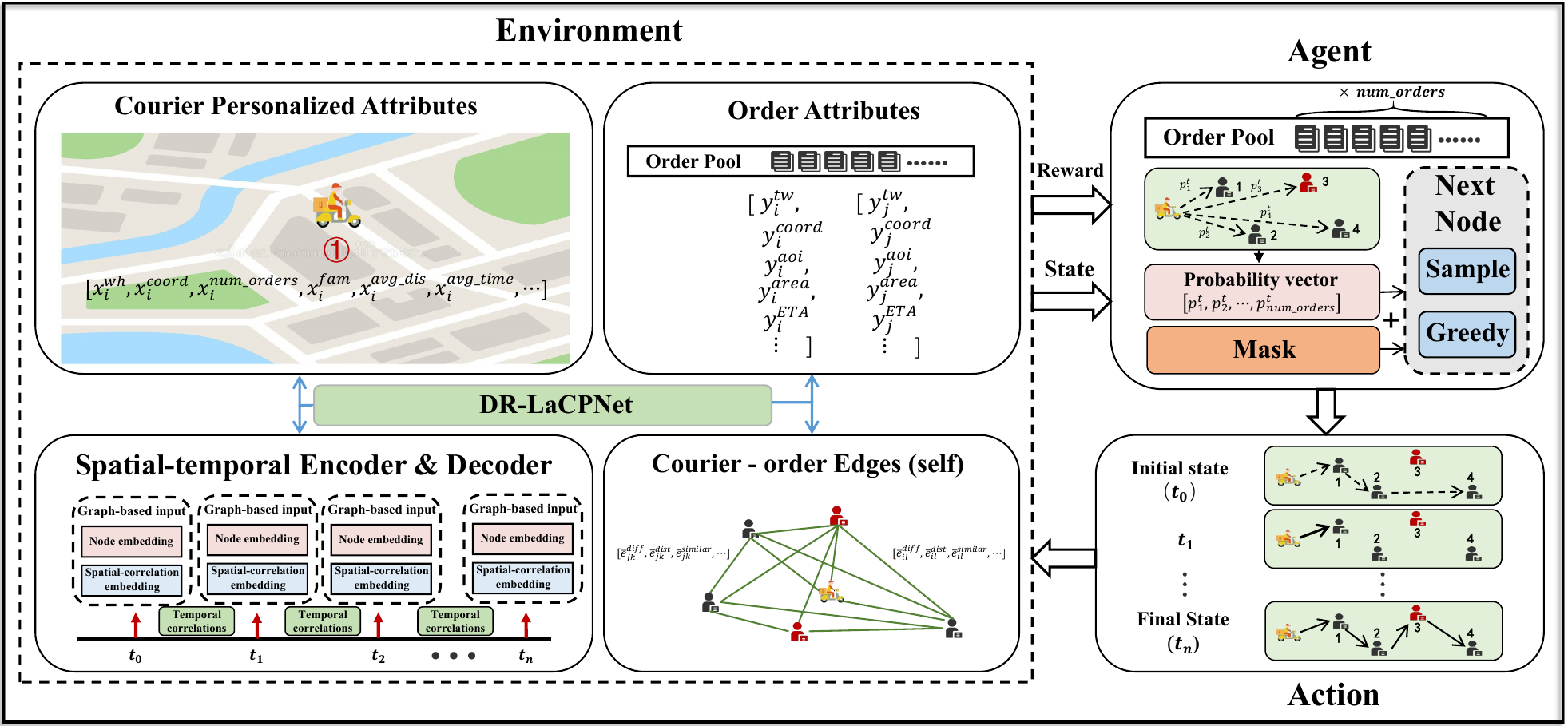}
	\caption{Reinforcement learning framework for the routing subproblem}
	\label{rrll}
\end{figure}

The reinforcement learning framework for the routing subproblem is illustrated in Figure \ref{rrll}, where the agent sequentially selects the next node and the environment transitions to a new state. After a small-scale preliminary experiments, we find that even for instances with only a few orders, commercial solvers often fail to obtain near-optimal solutions within the decision time budget $\Delta t$. Since each dispatching wave must be completed in real time, the computation time per wave is restricted to at most $\Delta t$.
Therefore, in order to reflect the goodness of solutions obtained by the DL model, we use the interactive update method of policy and baseline as our reinforcement learning algorithm, as detailed in Algorithm \ref{reinforce_lower}. At epoch 0, the baseline is initialized as a copy of the policy, which is then continuously updated using a greedy strategy. In order to further expand the solution space, the policy is updated using a more ``moderate'' sampling strategy instead of a greedy strategy. Here we use Adam as the optimizer to update $\theta$ \citep{kingma2015adam}. Also, a warm start with pre-trained weights is employed in reinforcement learning to get $\theta_{0}$, which accelerates the convergence and enhances training stability by leveraging prior knowledge. Unlike \citet{kool2018attention} and \citet{chen2025dynamic} directly replacing the baseline with the updated parameters when the difference between the solution obtained by policy and baseline is too large by t-test ($\overline{\alpha}=5\%$), we softly update the baseline, i.e., $\theta^{BL}_{\tau} = \mu \theta^{BL}_{\tau-1} + \lambda \theta_{\tau}$, where $\mu$ and $\lambda$ are the soft update factors, $\theta^{BL}_{\tau-1}$ is the coefficient of baseline in the previous epoch, and $\theta_{\tau}$ is the coefficient of policy in the current epoch. This weighted rather than direct substitution update prevents model oscillations and overfitting \citep{nikishin2018improving}, which improves the generalization ability of the model. 

\begin{algorithm}[ht]
	\renewcommand{\algorithmicrequire}{\textbf{Input:}}
	\renewcommand{\algorithmicensure}{\textbf{Output:}}
	\caption{Reinforcement Learning Algorithm for the routing subproblem}
	\label{reinforce_lower}
	\begin{algorithmic}[1]
    \STATE Input: number of epochs $\tau_{\max}$, batch size $B$, soft update factors $\mu, \lambda$, datasets $D^{Train}, D^{Valid}$
    \STATE Initialize: Load pre-trained weights for policy $\theta \leftarrow \theta_0$ (warm start), set baseline $\theta^{BL} \leftarrow \theta_0$
    \FOR{$\tau = 1, \dots, \tau_{\max}$}
    \FOR{$batch = 1, \dots, B$}
    \STATE \textbf{\textit{Sample}}: $s_i \leftarrow \text{RandomInstance}(D^{Train})$
    \STATE \textbf{\textit{Rollout}}: $\pi_i \leftarrow \text{SampleRollout}(s_i, \theta_{\tau})$, \quad $\pi_i^{BL} \leftarrow \text{GreedyRollout}(s_i, \theta_{\tau}^{BL})$
    \STATE \textbf{\textit{Update}}: $\theta_{\tau} \leftarrow \text{Adam}\left(\theta_{\tau} ,  \sum_{i=1}^B (L(\pi_i) - L(\pi_i^{BL})) \nabla_{\theta} \log p_{\theta}(\pi_i)\right)$
    \ENDFOR
    \STATE Validate the results on $D^{Valid}$
    \STATE Calculate $\overline{\alpha} = \text{t-test}\left(\{L(\pi_i)\}_{i=1}^B, \{L(\pi_i^{BL})\}_{i=1}^B \vert D^{Valid} \right)$
    \IF{significant difference ($\overline{\alpha} \leq 5\%$)}
    \STATE \textbf{\textit{Soft update}}: $\theta^{BL}_{\tau} = \mu \theta^{BL}_{\tau-1} + \lambda \theta_{\tau}$
    \ENDIF
    \ENDFOR
    \STATE Output: The parameters of the routing DL $\theta$
    \end{algorithmic}
\end{algorithm}

\subsubsection{DR-LaCPNet framework}
\begin{figure}[htb!]
	\centering
	\includegraphics[scale=0.5]{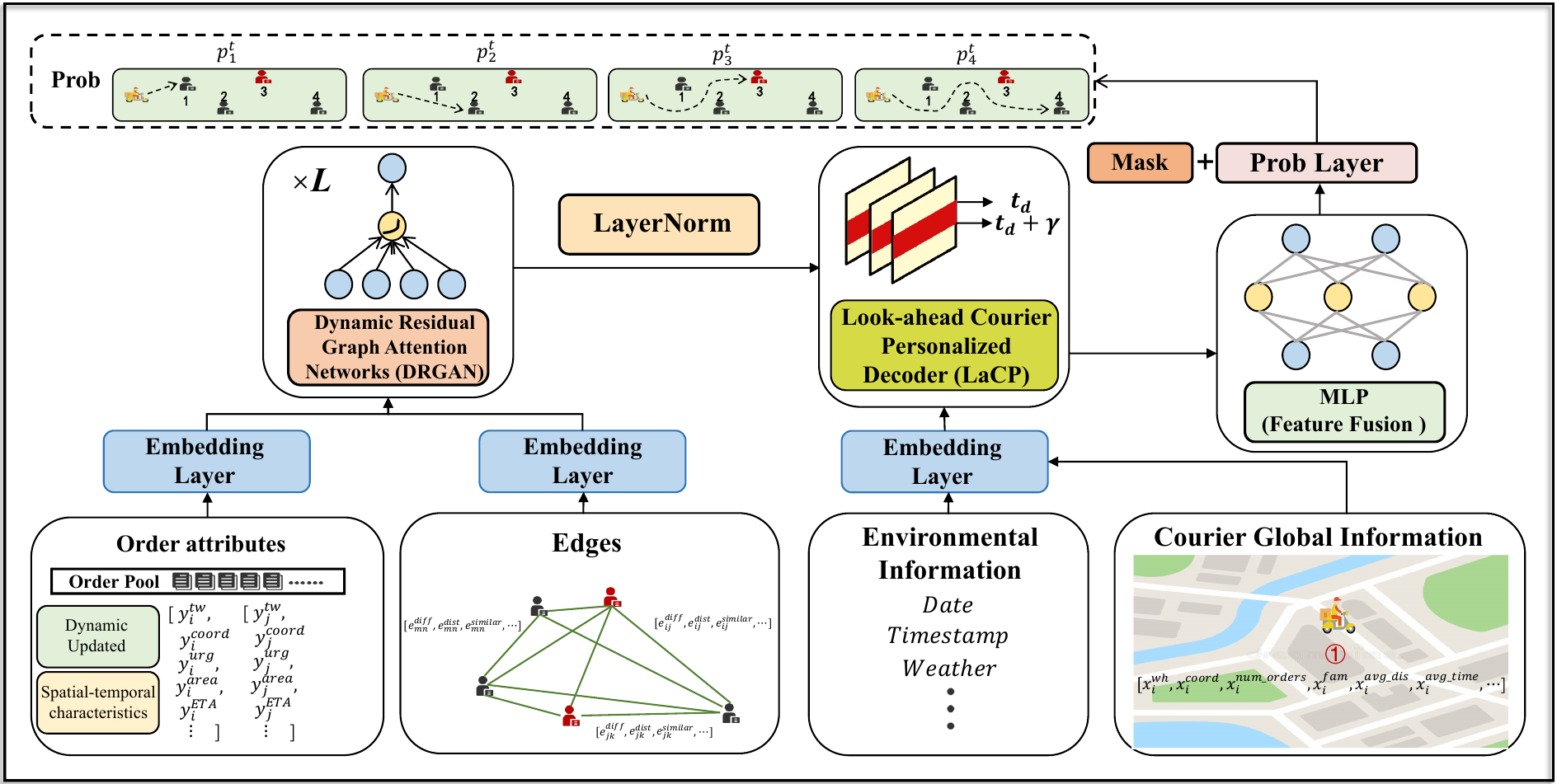}
	\caption{LaCPNet model framework}
	\label{ldl}
\end{figure}

The total framework of the proposed DR-LaCPNet for the routing subproblem is shown in Figure \ref{ldl}. We next describe the critical components in detail as follows.

\subsubsubsection{\textit{Input spatial-temporal graph structures}}

We use a spatial-temporal graph since travel times are time-dependent and each order is associated with temporal attributes (e.g., accept/promise times and time windows).
We integrate the spatial-temporal features into the subsequent module designed for feature enhancement, rather than using them as raw inputs without feature transformation. For each decision step $d$, $\mathcal{V}_d$ denotes the node set, $\mathcal{A}_d$ denotes the edge set, $\mathbf{X}_d$ denotes the node features, and $\mathbf{E}_d$ denotes the edge features. Given the initial information, we can construct the input spatial-temporal graph $\mathcal{G}_d^k=\left(\mathcal{V}_d, \mathcal{A}_d, \mathbf{X}_d, \mathbf{E}_d\right)$ to capture the spatial-temporal correlations in the routing subproblem. 
By padding the starting position of the courier to match the length of other nodes, we ensure all nodes have feature vectors of the same length. Then, the feature vector of node $i$, denoted by $\mathbf{x}_i$, is defined as:
\begin{equation}
\mathbf{x}_i=\left(\overline{x}_i, \overline{y}_i, X_i^{ac}, X_i^{pr}, a_i, b_i, AOI_{i}, I^{cur}_i, TW_i^{ur}, TW_i^{re}\right),
\label{c31}
\end{equation}
where $(\overline{x}_i, \overline{y}_i)$ are the coordinates, $X_i^{ac}$ is the accept time, $X_i^{pr}$ is the promised arrival time, $[a_i,b_i]$ is the time window, $AOI_{i}$ is the area of interests (\textit{AOI}) type and $I^{cur}_i$ is the indicator that whether node $i$ is the current courier position. To better distinguish the temporal difference of orders, we introduce $TW_i^{ur}=b_i-r$ as the time urgency and $TW_i^{re}= r-a_i$ as the time readiness, where $r$ represent the current time. The dimension of $\mathbf{x}_i$ is denoted as $D_{\mathbf{x}}$.

In addition, the edge features of the complete graph $\mathcal{G}_{d}^k$ is denoted as $e_{d i j}$, which constitutes the edge-feature tensor $\mathbf{E}_{d}$. Instead of directly feeding the raw features into subsequent modules, we slightly preprocess the edge features to reflect the temporal characteristics. Given the time-dependent travel time of each edge, we set $e_{d i j}= \left( t_{r_{d}^{k} i j}, \widetilde{A}_{d i j}, \delta_{d i j}^{\text {early}}, \delta_{d i j}^{\text {late}}, B_{d i j} \right)$. Here we consider the travel-time attributes from node $i$ to all remaining nodes when constructing edge features, where 
$r_{d}^{k}$ denotes the current time at step $d$. Specifically, we use $\widetilde{A}_{d i j}$ for estimated arrival time of node $j$, $\delta_{d i j}^{\text {early}}$ and $ \delta_{d i j}^{\text {late}}$ indicating whether arriving early of late at node $j$, and $B_{d i j}$ for time window penalties. The dimension of $e_{d i j}$ is denoted as $D_{e}$. 

To reduce the value discrepancy between node features $\mathbf{X}_d$ and edge features $\mathbf{E}_d$, we embed them through linear transformation layers separately. 
Let $D_h$ denote the dimension of the hidden layers.
For simplicity, we drop the subscript $d$ and the superscript $k$ when there is no ambiguity. 
The transformed node features $\overline{\mathbf{x}}_i$ and edge features $\overline{\mathbf{e}}_{i j}$ can be computed as:
\begin{equation}
\begin{aligned}
\overline{\mathbf{x}}_i & =\operatorname{ReLU} \left(\mathbf{W}_\mathbf{x} \mathbf{x}_i+\mathbf{b}_\mathbf{x} \right), \\
\overline{\mathbf{e}}_{i j} & =\operatorname{ReLU}  \left(\mathbf{W}_\mathbf{e} \mathbf{e}_{i j}+\mathbf{b}_\mathbf{e}\right),
\end{aligned}
\label{c32}
\end{equation}
where $\mathbf{W}_\mathbf{x} \in \mathbb{R}^{{D_h} \times D_{\mathbf{x}}}, \mathbf{b}_\mathbf{x} \in \mathbb{R}^{D_h}, \mathbf{W}_\mathbf{e} \in \mathbb{R}^{{D_h} \times D_{\mathbf{e}}}, \mathbf{b}_\mathbf{e} \in \mathbb{R}^{D_h}$ are trainable parameters. From this we obtain $\overline{\mathbf{x}}_i \in \mathbb{R}^{D_h}$ and $\overline{\mathbf{e}}_{i j} \in \mathbb{R}^{D_h}$. After processing, our input spatial-temporal graph can be easily fed into the subsequent dynamic residual graph attention network encoder.

\subsubsubsection{\textit{Dynamic Residual Graph Attention Network Encoder}}

In this section, we present the Dynamic Residual Graph Attention Network (DRGAN) Encoder, the crucial component of our proposed model designed to effectively capture the complex spatial-temporal relationships in the routing subproblem. This encoder leverages the strengths of Graph Attention Networks (GAT) and residual connections to enhance the representation of node and edge features, thereby improving the model's ability to make informed decisions.

Each GAT residual block consists of a GAT layer followed by an activation function, a residual connection, and a normalization layer. The GAT layer incorporates edge features into the node representation and then updates it. 
Here we set the initial input embeddings as the 0-th layer, i.e.,
$\mathbf{h}_i^0=\overline{\mathbf{x}}_i$.
The attention coefficients $z_{i,j}^{l}$ are then computed by Equation \eqref{c34} at each layer.
Since each edge $(i, j) \in \mathcal{A}$ contains multidimensional features, we first transform $\overline{\mathbf{e}}_{ij}$ into a scalar edge attribute $\tilde e_{ij}^{\,l}\in\mathbb{R}$ used in the attention mechanism. At the first layer, $\tilde e_{ij}^{\,0}$ is obtained from $\overline{\mathbf{e}}_{ij}$ via a lightweight learnable mapping; in deeper layers, $\tilde e_{ij}^{\,l}$ is dynamically updated using the attention weights from the previous layer. We then update node representations via one-hop attention aggregation and update edge representations layer-wisely using $z_{i, j}^{l}$, i.e., we set $\tilde{e}_{i j}^{l+1}=z_{i j}^l$.
\begin{equation}
\overline{\mathbf{h}_i^{{l}}}=\sum_{j \in \mathcal{N}(i)} z^{l}_{i, j} \boldsymbol{\Theta}_t \mathbf{h}_j^{{l}}
\label{c33}
\end{equation}
\begin{equation}
z_{i, j}^{l}=\frac{\exp \left(\operatorname{LeakyReLU}\left(\mathbf{q}_s^{\top} \boldsymbol{\Theta}_s \mathbf{h}_i^{{l}}+\mathbf{q}_t^{\top} \boldsymbol{\Theta}_t \mathbf{h}_j^{{l}}+\mathbf{q}_z^{\top} \boldsymbol{\Theta}_z \tilde e_{i j}^{\,l}\right)\right)}{\sum_{k \in \mathcal{N}(i)} \exp \left(\operatorname{LeakyReLU}\left(\mathbf{q}_s^{\top} \boldsymbol{\Theta}_s \mathbf{h}_i^{{l}}+\mathbf{q}_t^{\top} \boldsymbol{\Theta}_t \mathbf{h}_k^{{l}}+\mathbf{q}_z^{\top} \boldsymbol{\Theta}_z \tilde e_{i k}^{\,l}\right)\right)}
\label{c34}
\end{equation}
where $\mathbf{q}_s^{\top}, \boldsymbol{\Theta}_s, \mathbf{q}_t^{\top}, \boldsymbol{\Theta}_t, \mathbf{q}_z^{\top}, \boldsymbol{\Theta}_z$ are all trainable parameters and $\mathcal{N}(i)$ is the one-hop set of node $i$. Consequently, we apply the following residual mechanism to mitigate gradient vanishing and a normalization layer to stabilize the training process.
\begin{equation}
\mathbf{h}_i^{(l+1)}=\operatorname{LayerNorm}\left(\operatorname{ReLU}\left(\overline{\mathbf{h}}_i^{{l}}\right)+\mathbf{h}_i^{{l}} \right)
\label{c35}
\end{equation}

Finally, the output of the DRGAN encoder includes node representations 
$\mathbf{H}^E= \big[(\mathbf{h}_i^E)^\top\big]_{i\in\mathcal{V}} \in \mathbb{R}^{|\mathcal{V}|\times D_h}$ 
and edge representations 
$\mathbf{A}^E = \big[\tilde e_{ij}^{\,E}\big]_{(i,j)\in\mathcal{A}} \in \mathbb{R}^{|\mathcal{A}|\times 1}$.
These outputs are treated as the input of the following Look-Ahead Courier-Personalized Decoder.

\subsubsubsection{\textit{Look-Ahead Courier- Personalized Decoder}}

Building upon the DRGAN encoder's output, which includes node representations $\mathbf{H}^{E}$ and edge representations $\mathbf{A}^{E}$, the Look-Ahead Courier-Personalized (LaCP) Decoder further incorporates environmental and courier-specific information to refine routing decision-making.

Initially, we embed the environmental information $\mathbb{E}_{d}^{k} \in \mathbb{R}^{D_\mathcal{A}}$ to $\overline{\mathbb{E}}_{d}^{k} \in \mathbb{R}^{D_h}$ through a multilayer perceptron (MLP). Subsequently, we expand the embedded environmental information to match the number of edges, i.e., $ \widetilde{\mathbb{E}}_{d}^{k} \in \mathbb{R}^{\lvert \mathcal{A} \rvert \times {D_h}}$. We apply the same strategy to the courier global information and transform the $\mathbb{G}^{k} \in \mathbb{R}^{D_\mathbb{G}}$ into $ \widetilde{\mathbb{G}}^{k} \in \mathbb{R}^{\lvert \mathcal{A} \rvert \times {D_h}}$. The edge representations are updated by concatenating the courier's global information and the environmental information.
\begin{equation}
\tilde{\mathbf{A}}^{E}=\operatorname{ReLU}\left(\left(\mathbf{A}^{E} \|  \widetilde{\mathbb{G}}^{k} \|\widetilde{\mathbb{E}}_{d}^{k} \right) \cdot W_A + b_A\right)
\end{equation}
where $W_A \in \mathbb{R}^{(2D_h+1) \times D_h}$ and $b_{A} \in \mathbb{R}^{D_h}$ are trainable parameters and $\|$ means concatenation. 
Let $i_c$ denote the index of the current node at decision step $d$.
Given the edge embedding matrix $\tilde{\mathbf{A}}^E \in \mathbb{R}^{|\mathcal{A}| \times D_h}$, we extract the rows corresponding to outgoing edges $(i_c,j)\in\mathcal{A}$ to obtain a node-aligned edge-context matrix $\tilde{\mathbf{A}}_{i_c}^E \in \mathbb{R}^{|\mathcal{V}| \times D_h}$, where the $j$-th row represents the edge embedding from the current node $i_c$ to node $j$.
 
Due to the discreteness of the time-dependent travel times, relying solely on the travel time of the current time step or incorporating travel times of all time steps is not ideal. The reasons are as follows: 1) Relying solely on the travel time of the current time step undermines its ability to anticipate future conditions. 2) Using travel times from past steps provides limited additional benefit once the current travel-time snapshot is available, and it increases the input size unnecessarily. 3) Incorporating travel times from distant future time steps can introduce noise, thereby perturbing the current decision-making process. Therefore, we propose the look-ahead mechanism to balance the model's need for the current decision step and the future decision step. We set the look-ahead time window to $\gamma$ to determine how far the model should look ahead, then denote the look-ahead travel time matrix as $\mathbf{F} \in \mathbb{R}^{\gamma \times \lvert \mathcal{V} \rvert \times{\lvert \mathcal{V} \rvert}}$. Given the current node $i_c \in \mathcal{V}$, we extract the travel times from the current node to all other nodes across these matrices and transpose the result to obtain $\widetilde{\mathbf{F}}$. 
\begin{equation}
\widetilde{\mathbf{F}}=\left[\mathbb{T}_d\left(i_c, \cdot\right), \mathbb{T}_{d+1}\left(i_c, \cdot\right), \ldots, \mathbb{T}_{d+\gamma-1}\left(i_c, \cdot\right)\right]^T \in \mathbb{R}^{|\mathcal{V}| \times \gamma}
\end{equation}

We encode $\widetilde{\mathbf{F}}$ using an MLP defined as follows:
\begin{equation}
\overline{\mathbf{F}} = \operatorname{ReLU}\left(\widetilde{\mathbf{F}} \cdot \mathbf{W}_{\widetilde{\mathbf{F}}}+ \mathbf{b}_{\widetilde{\mathbf{F}}}\right)
\end{equation}
where $\mathbf{W}_{\widetilde{\mathbf{F}}} \in \mathbb{R}^{\gamma \times D_{h}}$ and $\mathbf{b}_{\widetilde{\mathbf{F}}} \in \mathbb{R}^{D_{h}}$ are trainable parameters. The output of the MLP is denoted as $\overline{\mathbf{F}} \in \mathbb{R}^{\lvert \mathcal{V} \rvert \times D_h}$. By fusing the travel time matrices of future $\gamma$ steps, we can enhance the model's ability to anticipate near-future travel conditions. Moreover, to explicitly represent the current node $i_c$ and indicate its position within the graph, we extract the embedding corresponding to $i_c$ from $\mathbf{H}^E$, denoted as $H_{i_c}^{E}$. This embedding is then expanded by repetition to match the number of nodes $\lvert \mathcal{V} \rvert$, forming $H_{i_c}^{E} \in \mathbb{R}^{\lvert \mathcal{V} \rvert \times D_h}$. We further encode the current-step travel times $\mathbb{T}_d\left(i_c, \cdot\right)$ by a separate MLP, yielding $\overline{\mathbf{F}}_{i_c} \in \mathbb{R}^{|\mathcal{V}| \times D_h}$.

To integrate the diverse factors influencing the decision-making process, we employ a composite node scorer to compute the final node scores. The proposed node scorer is an MLP which concatenates $\mathbf{H}^{E},  \overline{\mathbf{F}}, \tilde{\mathbf{A}}_{i_c}^E, H_{i_c}^{E}, \overline{\mathbf{F}}_{i_c}$ to enhance the feature representation.
\begin{equation}
h_d:= \operatorname{NodeScorer}\left(\mathbf{H}^{E} \| \overline{\mathbf{F}}\| \tilde{\mathbf{A}}_{i_c}^E \| H_{i_c}^{E} \| \overline{\mathbf{F}}_{i_c} \right) 
\end{equation}
The $\operatorname{\textit{Mask}}$ mechanism is then applied to the node scores according to the visited mask ${M}_d$ to get the probabilities $p_\theta\left(X_d \mid S_d\right)$ while choosing the next node $X_d$.
\begin{equation}
\operatorname{\textit{Mask}}\left(h_d, \mathbf{M}_d\right)= \begin{cases}h_d[i] & \text { if } \mathbf{M}_d[i]=1 \\ -10^9 & \text { if } \mathbf{M}_d[i]=0\end{cases}
\end{equation}
\begin{equation}
p_\theta\left(X_d \mid S_d\right)=\operatorname{Softmax}\left(\operatorname{\textit{Mask}}\left(h_d, \mathbf{M}_d\right)\right)
\end{equation}

We next introduce the proposed heuristic for the order dispatching problem. 

\subsection{Heuristic for the order dispatching subproblem} \label{heuristic_upper}

As stated earlier, the resulting model in Section \ref{lower_model} is then embedded into the dispatching process to rapidly score candidate couriers via routing-aware marginal cost estimation. Below gives the specific description of the position-pool greedy dispatch heuristic. 

As shown in Algorithm~\ref{alg_greedy_ls_dispatch}, the proposed algorithm includes the position-pool greedy dispatch heuristic and a local-search refinement phase (PP-Greedy-LS) to further improve solution quality. Specifically, the PP-Greedy-LS heuristic first evaluates each courier's current routing cost $J_{\text{old}}^{k}$ under the trained routing policy $\pi^{R}$, and aggregates them to obtain the initial system cost $J_{\text{tot}}=\sum_{k\in\mathcal K}J_{\text{old}}^{k}$. For each courier $k$, the heuristic then constructs a position pool $S_k$ consisting of the courier's current position $l_k$ and the positions of its currently dispatched orders $O^{old}_k$, which serves as a compact spatial representation of the courier's workload. Next, the new orders $O^{new}$ are sorted in a non-descending order of urgency $b_i$, yielding $\hat O^{new}$, so that more urgent orders are processed first. For each order $i\in\hat O^{new}$, the heuristic restricts dispatch candidates to a local neighborhood by selecting $\mathcal K_Q(i)$, the $Q$-nearest couriers with the smallest distance $\min_{p\in S_k}\|p-l_i\|$ to the order position $l_i$. Among these candidates, the heuristic chooses $k^\star$ that minimizes the incremental routing cost $\Delta_{ik}$, defined as the difference between the rollout routing cost after tentatively inserting order $i$ into courier $k$'s workload and the courier's current cost $J_{\text{old}}^{k}$, i.e., $\Delta_{ik}=\text{GreedyRolloutCost}(\pi^{R},k\cup\{i\})-J_{\text{old}}^{k}$. If a feasible $k^\star$ exists, the dispatch decision is set to $y_i^{k^\star}=1$ (and $y_i^k=0$ for all $k\neq k^\star$), the position pool is updated as $S_{k^\star}\leftarrow S_{k^\star}\cup\{l_i\}$, and both $J_{\text{old}}^{k^\star}$ and $J_{\text{tot}}$ are updated by adding $\Delta_{ik^\star}$; otherwise, the order is skipped for next dispatching.

\begin{algorithm}[htbp]
	\renewcommand{\algorithmicrequire}{\textbf{Input:}}
	\renewcommand{\algorithmicensure}{\textbf{Output:}}
	\caption{Position-pool greedy dispatch with local search refinement (PP-Greedy-LS)}
	\label{alg_greedy_ls_dispatch}
	\begin{algorithmic}[1]
	\STATE Input: new orders $O^{new}$, couriers $\mathcal K$, trained routing policy $\pi^R$, candidate size $Q$, max iterations $I$, acceptance probability $\sigma$
	\STATE Compute $J_{\text{old}}^{k}$ for all $k\in\mathcal K$ using $\pi^R$; set $J_{\text{tot}} \leftarrow \sum_{k\in\mathcal K} J_{\text{old}}^{k}$
	\STATE For each $k\in\mathcal K$, construct a courier position pool $S_k$ from the positions of $O^{old}_k$ and $l_k$
	\STATE Sort orders in $O^{new}$ in non-descending order by urgency $b_i$ to obtain $\hat O^{new}$
	\FOR{each order $i \in \hat O^{new}$}
	\STATE Select the top-$Q$ candidate couriers with the smallest $\min_{p\in S_k}\|p-l_i\|$, denoted by $\mathcal K_Q(i)$
	\STATE $k^\star \leftarrow \arg\min_{k\in\mathcal K_Q(i)} \Delta_{ik}$, where $\Delta_{ik}=\text{GreedyRolloutCost}(\pi^{R}, k\cup\{i\})-J_{\text{old}}^{k}$
	\IF{$k^\star$ exists}
	\STATE Set $y_i^{k^\star} \leftarrow 1$ and $y_i^{k} \leftarrow 0$ for all $k\neq k^\star$
	\STATE Update $S_{k^\star} \leftarrow S_{k^\star}\cup\{l_i\}$
	\STATE $J_{\text{old}}^{k^\star} \leftarrow J_{\text{old}}^{k^\star} + \Delta_{ik^\star}$, $J_{\text{tot}} \leftarrow J_{\text{tot}} + \Delta_{ik^\star}$
    \ELSE
    \STATE Continue
	\ENDIF
	\ENDFOR
	\STATE Set current dispatch $Y \leftarrow \{y_i^k\}$ and current cost $J \leftarrow J_{\text{tot}}$
	\STATE Set best dispatch $Y^\star \leftarrow Y$ and best cost $J^\star \leftarrow J$
	\FOR{$t=1$ to $I$}
	\STATE Generate a neighborhood $\mathcal{N}(Y)$ by dispatch moves
	\STATE Select $Y' \leftarrow \arg\min_{Z\in \mathcal N(Y)} J(Z)$, where $J(Z)$ is evaluated incrementally by $\pi^R$
	\IF{$J(Y')<J$}
	\STATE $Y \leftarrow Y'$, $J \leftarrow J(Y')$
	\IF{$J<J^\star$}
	\STATE $Y^\star \leftarrow Y$, $J^\star \leftarrow J$
	\ENDIF
	\ELSE
    \STATE Generate random number $rnd \in[0,1]$
	\IF{$rnd <\sigma$}
	\STATE Select a random $Y'\in \mathcal N(Y)$ and set $Y \leftarrow Y'$, $J \leftarrow J(Y')$
	\ENDIF
	\ENDIF
	\ENDFOR
	\STATE Output: best dispatch $Y^\star$ and $J^\star$
	\end{algorithmic}
\end{algorithm}

After completing the position-pool greedy dis-
patch heuristic, the algorithm sets the resulting dispatch as the current solution $Y\leftarrow \{y_i^k\}$ with cost $J\leftarrow J_{\text{tot}}$, and initializes the best solution and objective by $(Y^\star,J^\star)\leftarrow (Y,J)$. The algorithm then performs a local search for at most $I$ iterations. In each iteration, a neighborhood $\mathcal N(Y)$ is generated by applying dispatch moves (e.g., relocating an order to a different courier). For each candidate solution $Z\in\mathcal N(Y)$, its total routing cost $J(Z)$ is evaluated incrementally using $\pi^{R}$, i.e., by recomputing only the affected couriers' rollout costs while keeping others unchanged. The algorithm selects the best neighbor $Y'=\arg\min_{Z\in\mathcal N(Y)}J(Z)$. If $J(Y')<J$, the move is accepted, then the current solution and the current objective are updated, i.e., $Y\leftarrow Y'$ and $J\leftarrow J(Y')$). The best objective $J^\star$ would be updated if $J<J^\star$. Otherwise, the algorithm accepts the inferior solution with probability $\sigma$ to encourage exploration and escape local optima. Finally, the algorithm outputs the best dispatch $Y^\star$ and its associated routing cost $J^\star$.

Unlike directly training an upper-layer neural network to make order dispatching decisions, the proposed algorithm provides a rather lightweight strategy to solve different scales of dispatching instances. 
This is particularly important in real-world operations, where the number of active couriers and incoming orders within a region is inherently time-varying and uncertain, fluctuating due to demand surges, courier online/offline dynamics, and spatial-temporal imbalances. 
By selecting the $Q$-nearest candidate couriers based on the proximity measure $\min_{p\in S_k}\|p-l_i\|$, our algorithm significantly reduces the number of route evaluations and avoids unreasonable dispatch decisions to couriers whose current workload footprint is far from the order position.
In the algorithm, the underlying routing DRL serves as an oracle for our dispatch heuristic, enabling rapid route evaluation.
This allows relatively accurate and low-cost assessment of dispatch strategies without iteratively solving MILPs. 
Moreover, our routing oracle incorporates time-dependent travel times, which is able to reflect near-future routing implications. 
This yields more informative marginal-cost estimates, since $\Delta_{ik}$ is evaluated by time-dependent rollouts and captures downstream routing impacts, mitigating myopic dispatches under time-varying traffic conditions.
Finally, Through the subsequent local-search phase with the inferior solution acceptance mechanism, the algorithm could escape local optima and improve robustness.

\section{Numerical study} \label{numerical}
In this section, we conduct experiments to verify the performance and efficiency of the proposed framework. For parameter settings, we apply a fixed learning rate of $\eta = 0.001$ in the reinforcement learning algorithm, with gradient clipping enforced via a maximum L2 norm of 1.0. The batch size is set to 128.
In the DR-LaCPNet model, we configure embedding dimensions $D = 128$, number of GAT layers $L=5$.
The MILP was solved directly by the Gurobi solver in Python. The numerical studies are conducted on a high-performance workstation equipped with an NVIDIA RTX 4090D GPU with 24 GB memory, and an Intel Xeon Platinum 8481C processor with 16 virtual cores.

\subsection{Dataset description} \label{train_dataset}
In this work, we use the LaDe-P dataset, which focuses on dynamic package pick-up processes. It contains over 10.6 million packages and 619k courier trajectories generated by about 21k couriers across five Chinese cities during a six-month period. Each record provides information on the courier, package, position, and time of the pick-up event, with GPS coordinates perturbed within 10 meters for privacy protection. The dataset exhibits rich spatial-temporal patterns: for example, more than 70\% of packages belong to type-1 AOIs, most packages are picked up within three hours, and couriers show noticeable differences in efficiency and working profiles. These large-scale, diverse, and dynamic features make LaDe-P a representative and challenging benchmark for research in last-mile logistics.

\begin{figure}[htb!]
	\centering
	\includegraphics[scale=0.5]{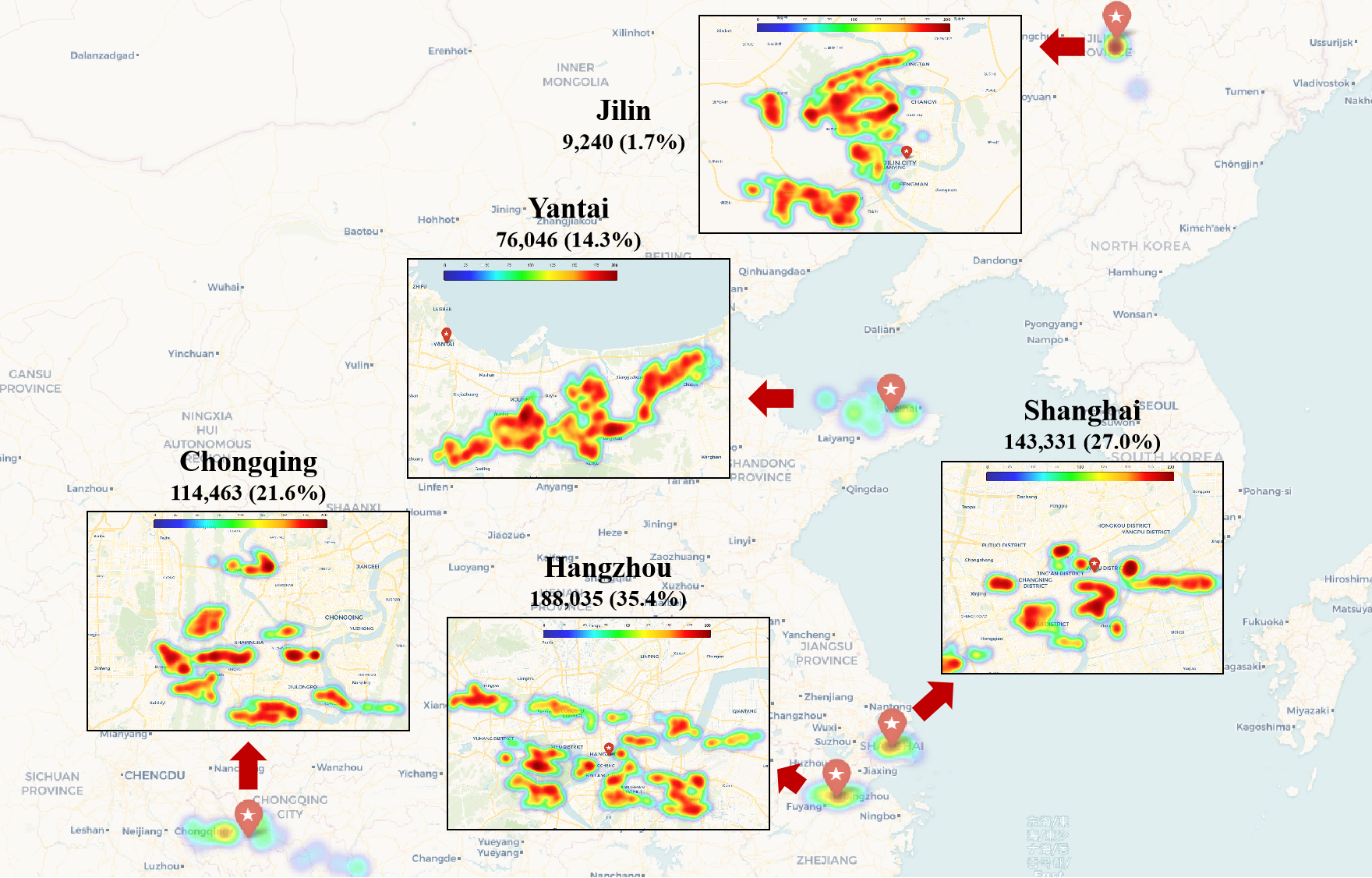}
	\caption{Spatial distribution across 5 cities}
	\label{map}
\end{figure}

Figure \ref{map} illustrates the spatial distribution across the five cities. It is evident that Hangzhou accounts for the largest share of the dataset. At the same time, different cities exhibit varying densities and cover distinct time periods, which provides us with diverse instances to validate the performance of our algorithms under heterogeneous urban environments and different experimental configurations.

\subsection{Dispatch instance simulator} \label{simulator}
Because the LaDe datasets are temporally sparse in each region, it is generally infeasible to directly extract a sufficiently large number of representative dispatching instances for scalable learning and testing. We therefore design a data-driven dispatch instance simulator that combines event replay with a \textsc{RollingUpdate} strategy to dynamically update the context, enabling the generation of abundant training instances while preserving the spatial--temporal patterns observed in the original data, as shown in Algorithm \ref{dispatch_simulator}. 

To construct dispatch instances for experiments and routing instances for training with controlled and diverse scales, we specify two scale parameters, i.e., the number of couriers $K$ and the number of new orders $N$, and vary them within $[K_{\min},K_{\max}]$ and $[N_{\min},N_{\max}]$ across instances. 
Firstly, $\textsc{InitContext}(\mathcal{D},\mathcal{R},t_0)$ initializes the context.
For each dispatching wave $h$, the simulator then draws $K$ candidate couriers from an empirical snapshot pool $\mathcal{P}(t_h,\mathcal{R})$, and retrieves each courier $k$'s in-hand orders $O_k^{\mathrm{old}}$ from the historical records. 
Next, the simulator performs event replay to collect a candidate set of order accept events $\tilde{O}$ within time $(t_h,t_h+\Delta]$ and region $\mathcal{R}$. 
To control the instance scale while maintaining the spatial-temporal characteristics of the raw dataset, we set each event $o\in\tilde{O}$ to be independently activated with probability 
$p_{\mathrm{act}}=\min\{1,\,N/|\tilde{O}|\}$ to yield the new order set $O^{new}$. If the realized $|O^{new}|$ exceeds $N$, we subsample $O^{new}$ to length $N$.

For each activated order, we construct its node features according to Equation~\eqref{c31} at time $t_h$. The resulting dispatch instance is built as $\mathcal{I}_h=(\mathcal{K},O^{new},t_h,\Delta,\Gamma_h)$ and is solved by the proposed dispatch procedure and routing rollouts under the pretrained routing policy $\pi^R$. After obtaining the dispatch decision $\{y_i^k\}$, the simulator advances to the next wave via \textsc{RollingUpdate} introduced in Algorithm~\ref{alg_state_transition}. Specifically, for each courier $k$, we merge its in-hand orders with the newly dispatched orders (i.e., $i\in O^{new}:y_i^k=1$), run $\pi^R$ to solve the routing subproblem and obtain arrival times $\{T_{k,i}\}_{i\in O_k^{\mathrm{cur}}}$. Subsequently, the old orders $O^{old}_{k}$ are updated, while the courier position is also updated as the last visited position before $t_h+\Delta$. 
Repeating this loop produces a temporally consistent, state-evolving sequence of dispatching waves.
Specifically, the decision at wave $h$ is propagated to wave $h{+}1$ through \textsc{RollingUpdate}, making subsequent instances and decisions conditional on earlier outcomes.
Together with the scale controls $(K,|O^{new}|)$, this provides a convenient way to generate continuous benchmark sequences spanning diverse instance sizes while preserving key spatial--temporal patterns in the raw logs.

\begin{algorithm}[ht]
\renewcommand{\algorithmicrequire}{\textbf{Input:}}
\renewcommand{\algorithmicensure}{\textbf{Output:}}
\caption{Dispatch Instance Simulator}
\begin{algorithmic}[1]
\STATE \textbf{Input:} Dataset $\mathcal{D}$, region $\mathcal{R}$, start time $t_0$, dispatching wave length $\Delta$, horizon $H$,
courier number range $[K_{\min},K_{\max}]$, order number range $[N_{\min},N_{\max}]$, routing policy $\pi^R$
\STATE \textbf{Output:} Dispatch instances $\{\mathcal{I}_h\}_{h=0}^{H-1}$

\STATE Initialize rolling context $\Gamma_0 \leftarrow \textsc{InitContext}(\mathcal{D},\mathcal{R},t_0)$
\FOR{$h=0,1,\dots,H-1$}
    \STATE Set dispatching wave time $t_h \leftarrow t_0 + h\Delta$
    \STATE Sample $K \sim \mathrm{Unif}(\{K_{\min},\ldots,K_{\max}\})$ and $N \sim \mathrm{Unif}(\{N_{\min},\ldots,N_{\max}\})$
    \STATE Sample $K$ couriers from the empirical snapshot pool $\mathcal{P}(t_h,\mathcal{R})$; obtain courier set $\mathcal{K}$ and, for each courier $k\in\mathcal{K}$, its in-hand orders $O_k^{\mathrm{old}}$    
    \STATE Collect candidate accept events $\tilde{O}$ within time $(t_h,t_h+\Delta]$ and region $\mathcal{R}$
    \STATE Set $p_{\mathrm{act}} \leftarrow \min\{1,\, N / |\tilde{O}|\}$
    \STATE Activate each $o\in\tilde{O}$ independently with probability $p_{\mathrm{act}}$ to obtain $O^{new}$
    \STATE If $|O^{new}|>N$, subsample $O^{new}$ to size $N$
    \STATE Construct node states $\{\mathbf{x}_i\}$ for all orders in $O^{new}$ according to Equation~\eqref{c31} at $t_h$
    \STATE Build instance $\mathcal{I}_h \leftarrow (\mathcal{K}, O^{new}, t_h, \Delta, \Gamma_h)$
    \STATE Solve $\mathcal{I}_h$ to obtain dispatch decision $\{y_i^k\}$ and routing rollouts under $\pi^R$
    \STATE $\Gamma_{h+1}\leftarrow \textsc{RollingUpdate}(\Gamma_h,\mathcal{I}_h,\{y_i^k\},\pi^R,t_h,\Delta)$ according to Algorithm~\ref{alg_state_transition}
\ENDFOR
\STATE \textbf{Return} $\{\mathcal{I}_h\}_{h=0}^{H-1}$
\end{algorithmic}
\label{dispatch_simulator}
\end{algorithm}

\begin{algorithm}[ht]
\renewcommand{\algorithmicrequire}{\textbf{Input:}}
\renewcommand{\algorithmicensure}{\textbf{Output:}}
\caption{\textsc{RollingUpdate}}
\label{alg_state_transition}
\begin{algorithmic}[1]
\STATE \textbf{Input:} Current rolling context $\Gamma_h$, instance $\mathcal{I}_h=(\mathcal{K},O^{new},t_h,\Delta,\Gamma_h)$, dispatch $\{y_i^k\}$, routing policy $\pi^R$, wave time $t_h$, wave length $\Delta$
\STATE \textbf{Output:} Next rolling context $\Gamma_{h+1}$

\FOR{each courier $k\in\mathcal{K}$}
    \STATE Let $O_k^{\mathrm{cur}} \leftarrow O_k^{\mathrm{old}} \cup \{i\in O^{new}:y_i^k=1\}$
    \STATE Run $\pi^R$ on $O_k^{\mathrm{cur}}$ starting at time $t_h$ to solve the routing subproblem and obtain arrival times $\{T_{k,i}\}_{i\in O_k^{\mathrm{cur}}}$
    \STATE Set $O_k^{\mathrm{old}} \leftarrow \{i\in O_k^{\mathrm{cur}}: T_{k,i}>t_h+\Delta\}$
    \STATE Update courier position to the last visited position before $t_h+\Delta$
\ENDFOR

\STATE Update time in the rolling context to $t_h+\Delta$ and return the updated courier positions and carry-over order sets as $\Gamma_{h+1}$
\STATE \textbf{Return} $\Gamma_{h+1}$
\end{algorithmic}
\end{algorithm}

\subsection{Learning performance} \label{learning_perform}
This section discusses the parameter settings and the learning curve of the LaCPNet.
We train our LaCPNet for 100 epochs with a batch size of 64, a learning rate of $lr=10^{-3}$, hidden dimension of 64, and a look-ahead step of 3. 
The objective weights are set to be $\alpha=0.7$ and $\phi=1.0$ by default.

\begin{figure}[htbp]
	\centering
	\begin{subfigure}{0.49\linewidth}
		\centering
		\includegraphics[width=1\linewidth]{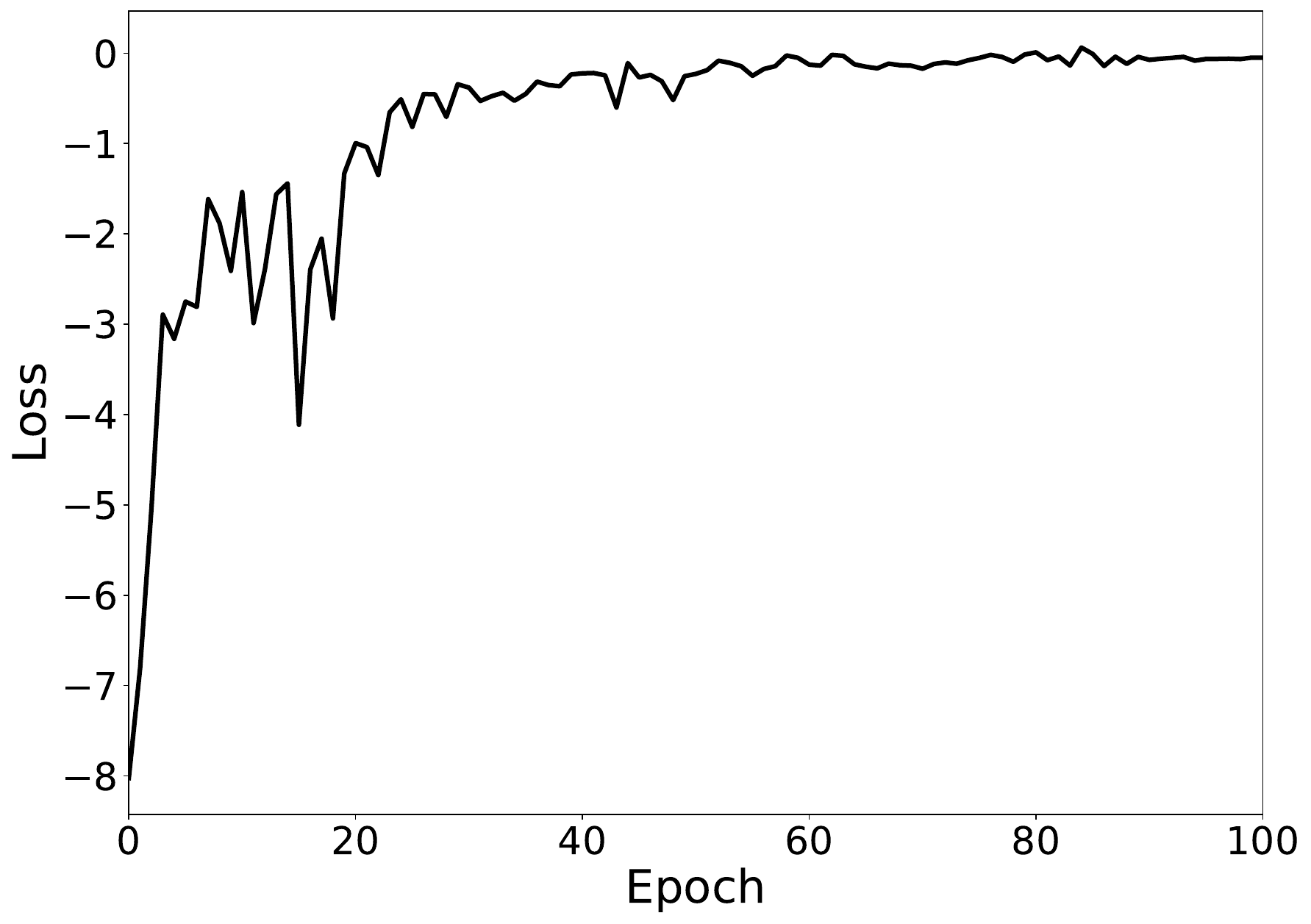}
		\caption{Training loss with epochs}
		\label{training_loss}
	\end{subfigure}
	\centering
	\begin{subfigure}{0.49\linewidth}
		\centering
		\includegraphics[width=1\linewidth]{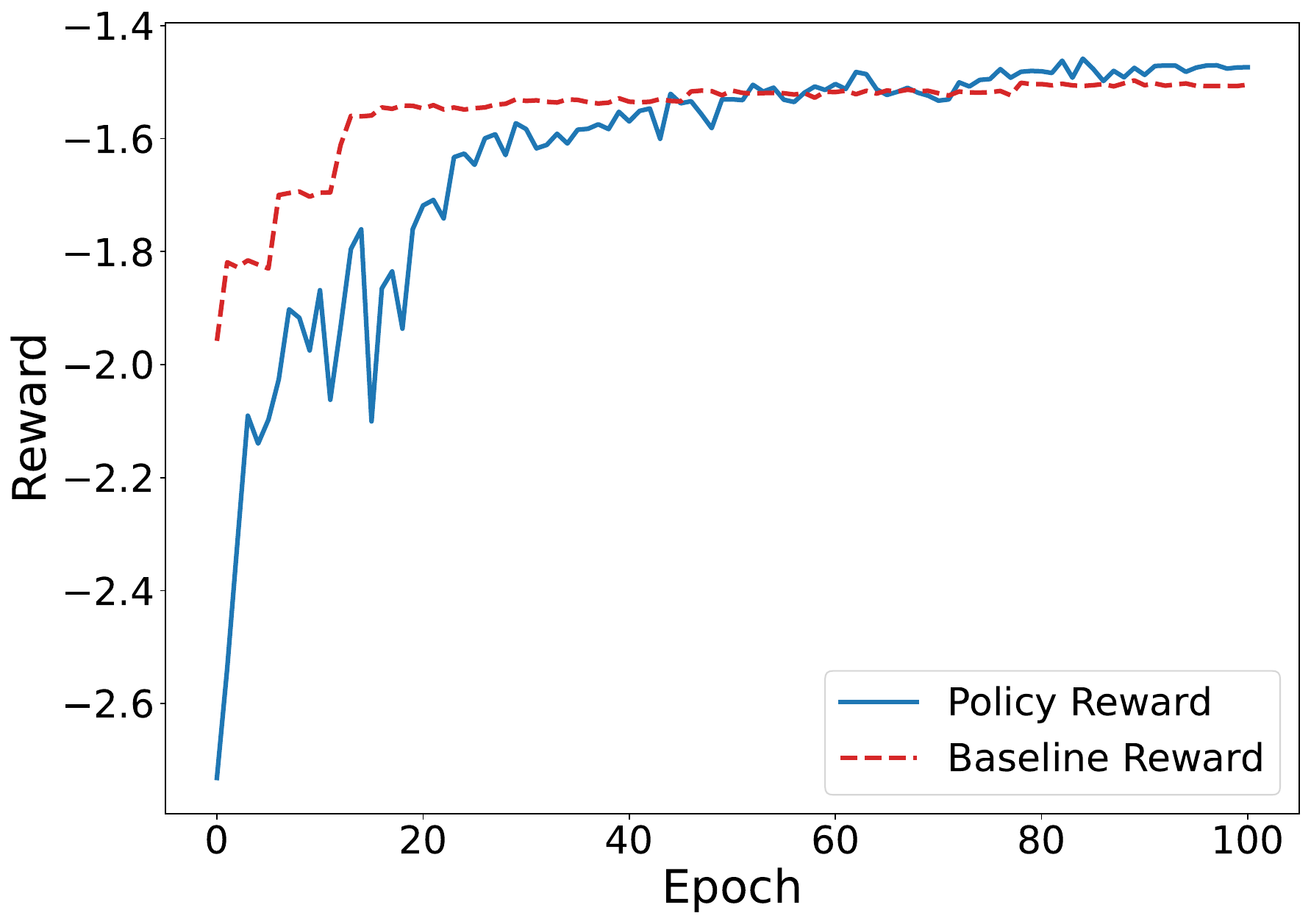}
		\caption{Training rewards of policy and baseline}
		\label{training_rewards}
	\end{subfigure}
	\caption{Learning curve of DR-LaCPNet}
	\label{tourt}
\end{figure}

During training process, the policy $\pi_i$ is evaluated against its baseline counterpart $\pi_i^{BL}$, and the objective is defined as the difference between their respective losses, i.e., $\mathcal{L}_{\pi_i}-\mathcal{L}_{\pi_i^{BL}}$. 
As shown by the convergence curves in Figure ~\ref{training_loss}, the loss decreases steadily until iteration 50. The loss then stabilizes and further training provides only marginal improvements at the cost of significant computational overhead.
Consistently, Figure~\ref{training_rewards} reports the episodic rewards of $\pi_i$ and $\pi_i^{BL}$, where both curves improve sharply within the first $\sim 50$ epochs and converge thereafter, with the policy reward gradually catching up to the baseline, while further gains become marginal. 
Consequently, for all subsequent evaluations, we adopt the model obtained at iteration 50 as the representative result. 

\subsection{Numerical experiments} \label{exp}
\subsubsection{Evaluation of the DRL routing oracle} \label{exp_1}
To ensure both the computational efficiency and decision quality of the upper-level dispatching heuristic, the solution quality and solution time of the DRL routing policy is of significance.
Therefore, we first evaluate the effectiveness of the proposed DRL routing oracle across different number of new orders $|O_k^{\mathrm{new}}|$ for a specific courier, including its in-hand orders and newly dispatched orders. 
We begin with a benchmark against the MILP formulation solved by commercial solver Gurobi. 
The following approaches are used as baselines for comparison:
1) MILP: The Mixed-Integer Linear Programming (MILP) of TDTSPTW-PU is solved using the Gurobi solver.
2) Greedy: A simple heuristic generating the routes by a step-by-step policy, selecting the next node with the shortest travel time at each step.
3) TS: A tabu search algorithm proposed by \citet{brandao2004tabu} for the open VRP. To apply it to TDTSPTW-PU, we specialize it to the single-vehicle ase and use 2-opt exchanges as the neighborhood structure.
4) DeepRoute: A deep model with a Transformer encoder and an attention-based decoder to rank unserved orders proposed by \citet{wen2021package}.
5) Graph2Route: A deep model proposed by \citet{wen2022graph2route}, which formulates the task as a graph-to-sequence problem. It applies a GNN to learn task embeddings and a sequence decoder to generate the route, effectively modeling spatial correlations among orders.
The maximum solution time of Gurobi is set to 300s, since dispatch decisions must be computed within each short dispatching wave to remain actionable.

\begin{table*}[htbp]
\centering
\caption{Performance comparison for DRL routing oracle}
\resizebox{\textwidth}{!}{%
\begin{tabular}{cc ccc cc cc cc cc ccc}
\toprule
\multicolumn{2}{c}{Scenario} 
& \multicolumn{3}{c}{MILP}
& \multicolumn{2}{c}{Greedy}
& \multicolumn{2}{c}{TS}
& \multicolumn{2}{c}{Deeproute}
& \multicolumn{2}{c}{Graph2Route}
& \multicolumn{3}{c}{DR-LaCPNet } \\
\cmidrule(lr){1-2}\cmidrule(lr){3-5}\cmidrule(lr){6-7}\cmidrule(lr){8-9}\cmidrule(lr){10-11}\cmidrule(lr){12-13}\cmidrule(lr){14-16}
$|O_k^{\mathrm{new}}|$ & $\delta$
& $UB$ & $Time$ & $Gap$
& $Obj$ & $Time$
& $Obj$ & $Time$
& $Obj$ & $Time$
& $Obj$ & $Time$
& $Obj$ & $Time$ & $Obj \; Imp.$\\
\midrule

\multirow{3}{*}{$[1,4]$}
& 1 & 0.50 & 34.26 & 0.00 & 0.65 & 0.02 & 0.50 & 0.76 & 0.50 & 0.01 & 0.50 & 0.02 & \textbf{0.50} & 0.03 & 0.00 \\
& 3 & 0.31 & 41.55 & 0.00 & 0.42 & 0.01 & 0.31 & 0.32 & 0.31 & 0.02 & 0.31 & 0.01 & \textbf{0.31} & 0.02 & 0.00 \\
& 5 & 0.19 & 31.08 & 0.00 & 0.28 & 0.03 & 0.19 & 0.55 & 0.19 & 0.01 & 0.19 & 0.03 & \textbf{0.19} & 0.04 & 0.00 \\
\addlinespace

\multirow{3}{*}{$[5,8]$}
& 1 & 1.17 & 120.45 & 0.00 & 1.55 & 0.05 & 1.25 & 25.63 & 1.28 & 0.04 & 1.22 & 0.05 & \textbf{1.17} & 0.09 & 0.00 \\
& 3 & 1.36 & 157.26 & 0.00 & 1.85 & 0.04 & 1.48 & 41.97 & 1.52 & 0.06 & 1.45 & 0.04 & \textbf{1.36} & 0.13 & 0.00 \\
& 5 & 1.28 & 300.00 & 24.7 & 1.45 & 0.07 & 1.16 & 33.55 & 1.20 & 0.05 & 1.18 & 0.06 & \textbf{1.11} & 0.08 & 4.31 \\
\addlinespace

\multirow{3}{*}{$[9,12]$}
& 1 & -- & -- & -- & 2.10 & 0.11 & 1.74 & 55.05 & 1.76 & 0.08 & 1.71 & 0.09 & \textbf{1.52} & 0.21 & 11.11 \\
& 3 & -- & -- & -- & 2.05 & 0.09 & 1.68 & 64.33 & 1.65 & 0.09 & 1.62 & 0.12 & \textbf{1.41} & 0.26 & 12.96 \\
& 5 & -- & -- & -- & 2.35 & 0.12 & 1.85 & 52.45 & 1.82 & 0.07 & 1.78 & 0.10 & \textbf{1.55} & 0.19 & 12.92 \\
\addlinespace

\multirow{3}{*}{$[13,16]$}
& 1 & -- & -- & -- & 4.50 & 0.18 & 3.85 & 98.04 & 3.75 & 0.13 & 3.62 & 0.15 & \textbf{3.15} & 0.29 & 12.98 \\
& 3 & -- & -- & -- & 4.20 & 0.15 & 3.35 & 103.82 & 3.25 & 0.16 & 3.03 & 0.14 & \textbf{2.72} & 0.34 & 10.23 \\
& 5 & -- & -- & -- & 3.50 & 0.22 & 2.65 & 96.89 & 2.60 & 0.14 & 2.50 & 0.18 & \textbf{2.12} & 0.27 & 15.20 \\

\bottomrule
\end{tabular}%
}
\label{compare_oracle}
\end{table*}

The results are demonstrated in Table \ref{compare_oracle}, where columns ``Obj'' and ``Time'' reports the objective value and solution time of the methods, respectively.  
For MILP, $UB$ denotes Gurobi's best incumbent objective and $Gap$ is the average optimality gap, i.e.,  $\textit{Gap}=(UB-LB)/UB\times 100 \% $, where $LB$ is Gurobi's lower bound. $Obj\;Imp.$ reports the objective improvement of DR-LaCPNet relative to the best objective value achieved by all methods (including DR-LaCPNet). Thus, $Obj\,Imp.=0$ if DR-LaCPNet attains the best objective.
It can be observed that on small- and medium-scale route instances, e.g., $|O_k^{\mathrm{new}}|\in[1,4]$ and $|O_k^{\mathrm{new}}|\in[5,8]$, the proposed DR-LaCPNet attains objective values that are competitive with the MILP baseline while requiring substantially shorter computation times.
On larger instances, e.g., $|O_k^{\mathrm{new}}|\in[9,12]$ and $|O_k^{\mathrm{new}}|\in[13,16]$, where Gurobi frequently fails to identify a feasible solution within the prescribed time limit (``--''), DR-LaCPNet continues to generate competitive results compared with other baselines. 
Specifically, compared with classical construction heuristics (Greedy and TS), LaCPNet achieves markedly better or comparable objective values across all instances while retaining sub-second runtime, making it suitable for time-sensitive routing evaluations. 
Compared with other deep models (DeepRoute and Graph2Route), LaCPNet consistently attains lower objective values across most scenarios, suggesting the benefit of jointly leveraging DRGAN encoder and LaCP decoder.
It is also noteworthy that the solution time of LaCPNet does not increase significantly with larger instance sizes, but instead remains relatively stable at less than 0.1 seconds. 
This confirms that our LaCPNet can fully account for environmental factors within an extremely short timeframe, thereby delivering high-quality solutions and providing a robust solution oracle for order dispatching heuristics.

\subsubsection{Offline evaluation} \label{exp_2}
To investigate the performance of our proposed integrated optimization framework for order dispatch and route planning, we first conduct offline evaluations across different cities in the LaDe dataset to isolate solution quality and runtime under controlled settings. As the experimental results shown in Section \ref{exp_1}, DR-LaCPNet consistently delivers superior solutions compared with other baselines, we therefore use the trained DR-LaCPNet as a unified routing oracle ($\pi^R$) for all dispatching methods when computing objectives for ${y_i^k}$. As a results, the reported objective values reflect only the quality of the dispatching policy rather than differences in routing evaluation

For this offline evaluation, we construct dispatch instances from the LaDe dataset across five cities. Each instance consists of $N^{\mathrm{new}}$ new orders and $K$ candidate courier. 
We benchmark PP-Greedy-LS against four baselines: 
(i) DRL Marginal-cost Assignment (DMCA), which solves a static dispatching problem by minimizing the DRL-estimated marginal routing cost. Specifically, for each courier $k$, we first compute its base routing cost under the DRL oracle, and then evaluate the incremental cost of assigning each new order $i$ to $k$ via a one-order rollout. These marginal costs form a cost matrix, based on which we solve a minimum-cost bipartite assignment using Gurobi (with an optional per-courier capacity constraint), yielding the dispatch decision ${y_i^k}$; 
(ii) Greedy (Algorithm~\ref{alg_append1}), which processes orders in non-descending urgency and assigns each order to the courier that minimizes the routing-cost increment within a local candidate set screened by current-position distance;
(iii) PP-Greedy (Algorithm~\ref{alg_append2}), which replaces the screening rule by a position-pool distance that accounts for proximity to any already-committed stop; and 
(iv) TS-Dispatching, which generates solutions using a tabu search that repeatedly relocates a newly dispatched order between couriers. Each candidate move is scored by recomputing the induced cost with the trained DR-LaCPNet, and a tabu list with aspiration is used to avoid cycling.

\begin{table*}[htbp]
\centering
\caption{Performance comparison for dispatching heuristics}
\resizebox{\textwidth}{!}{%
\begin{tabular}{lcc cc cc cc cc ccc}
\toprule
\multicolumn{3}{c}{Scenario}
& \multicolumn{2}{c}{DMCA}
& \multicolumn{2}{c}{Greedy}
& \multicolumn{2}{c}{PP-Greedy}
& \multicolumn{2}{c}{TS-Dispatching}
& \multicolumn{3}{c}{PP-Greedy-LS} \\
\cmidrule(lr){1-3}
\cmidrule(lr){4-5}\cmidrule(lr){6-7}\cmidrule(lr){8-9}\cmidrule(lr){10-11}\cmidrule(lr){12-14}
City & $|O^{\mathrm{new}}|$ & $K$
& Obj & Time
& Obj & Time
& Obj & Time
& Obj & Time
& Obj & Time & Obj Imp. \\
\midrule

\multirow{3}{*}{Shanghai}
& 8  & 5  & \textbf{2.30} & 1.51 & 3.43 & 0.38 & 3.43 & 0.38 & \textbf{2.30} & 22.52 & \textbf{2.30} & 1.07  & 0.00 \\
& 20 & 9  & 10.38 & 6.87 & 6.17 & 2.59 & 6.17 & 2.61 & \textbf{6.11} & 89.29 & \textbf{6.11} & 40.91 & 0.00 \\
& 27 & 13 & 10.36 & 18.47 & 8.91 & 4.37 & 8.77 & 4.29 & 8.86 & 97.87 & \textbf{8.57} & 93.62 & 2.38 \\
\addlinespace

\multirow{3}{*}{Hangzhou}
& 8  & 5  & 4.28 & 0.60 & 4.67 & 0.56 & 4.67 & 0.56 & \textbf{4.20} & 22.20 & 4.28 & 1.02  & -1.94 \\
& 22 & 8  & 7.46 & 5.14 & 8.64 & 3.29 & 8.64 & 3.26 & 7.06 & 45.76 & \textbf{6.92} & 23.42 & 2.01 \\
& 24 & 12 & 10.66 & 7.26 & 12.52 & 3.27 & 12.52 & 3.28 & 10.17 & 67.61 & \textbf{8.78} & 30.32 & 13.68 \\
\addlinespace

\multirow{3}{*}{Chongqing}
& 9  & 6  & \textbf{14.65} & 0.57 & \textbf{14.65} & 0.52 & \textbf{14.65} & 0.51 & \textbf{14.65} & 21.90 & \textbf{14.65} & 0.97 & 0.00 \\
& 16 & 10 & 40.14 & 4.26 & 41.00 & 1.92 & 41.12 & 1.90 & \textbf{39.55} & 32.34 & 39.70 & 6.31 & -0.38 \\
& 24 & 12 & \textbf{30.60} & 8.96 & 36.99 & 2.79 & 34.84 & 2.69 & \textbf{30.60} & 73.36 & \textbf{30.60} & 25.86 & 0.00 \\
\addlinespace

\multirow{3}{*}{Jilin}
& 9  & 8  & \textbf{2.69} & 2.30 & 4.71 & 1.89 & 2.74 & 1.74 & \textbf{2.69} & 21.28 & \textbf{2.69} & 6.48 & 0.00 \\
& 19 & 11 & \textbf{28.44} & 6.20 & 28.50 & 2.95 & 28.50 & 2.97 & \textbf{28.44} & 67.32 & 28.50 & 21.66 & -0.22 \\
& 33 & 14 & 56.89 & 30.70 & 57.00 & 7.02 & 57.00 & 6.86 & \textbf{49.44} & 420.36 & 52.86 & 132.07 & -6.92 \\
\addlinespace

\multirow{3}{*}{Yantai}
& 8  & 6  & 26.31 & 1.28 & 26.79 & 0.97 & 26.79 & 0.96 & 23.49 & 7.57 & \textbf{23.34} & 3.19 & 0.66 \\
& 22 & 10 & 34.17 & 9.64 & 57.73 & 4.60 & 57.73 & 4.61 & \textbf{29.81} & 49.77 & 31.38 & 33.58 & -5.25 \\
& 28 & 13 & 35.80 & 10.09 & 63.01 & 5.84 & 65.80 & 4.51 & 32.78 & 39.18 & \textbf{32.60} & 44.76 & 0.58 \\
\bottomrule
\end{tabular}%
}
\label{compare_dispatch}
\end{table*}

Overall, Table~\ref{compare_dispatch} shows that PP-Greedy-LS delivers the best objectives in most settings (bold entries) with acceptable solution times, demonstrating the benefit of combining position-pool greedy construction with a lightweight local search. 
The advantage becomes more pronounced as the instance size grows: in Shanghai with $N^{\mathrm{new}}=27$, PP-Greedy-LS reduces the objective to 8.57 versus 8.77 for PP-Greedy; in Hangzhou, it achieves 6.92 versus 7.06 at $N^{\mathrm{new}}=22$ and 8.78 versus 10.17 at $N^{\mathrm{new}}=24$. 
Meanwhile, TS-Dispatching is consistently more time-consuming, e.g., 420.36s for Jilin's instance with $N^{\mathrm{new}}=33$ and $K=14$, whereas PP-Greedy-LS completes in 132.07s while maintaining competitive solution quality. 
These results support PP-Greedy-LS as an effective and efficient upper-level dispatching heuristic when paired with DR-LaCPNet for rapid routing evaluation.

\subsubsection{Online rolling-horizon simulation} \label{exp_3}
To further investigate the operational effectiveness and robustness of the proposed integrated optimization framework, we use the dispatch instance simulator introduced in Algorithm \ref{dispatch_simulator} to dynamically assess the cumulative system-level impact under repeated decision making within strict per-wave time budgets. Specifically, we consider the continuous dispatching results for each city within one hour (i.e., 12 dispatching waves) and calculate the average results for each dispatching wave.

The following metrics are used to evaluate the solution quality and stability of different methods:
1) \textbf{Avg. Obj:} the average objective value.
2) \textbf{Avg. TWVR:} the average time-window violation rate. 
For an order $i\in O^{new}$ dispatched to courier $k$, a violation occurs if $\beta_i^{k}<a_i$ or $\beta_i^{k}>b_i$, counting both early and late pickups.
3) \textbf{Avg. TWP:} the average time-window penalties defined in Equation \eqref{c19}. 
4) \textbf{Max. TWP:} the maximum realized time-window penalty, which reflects tail-risk service failures that may be obscured by averages.
5) \textbf{Avg. TT:} the average travel time per courier within a dispatching wave.
6) \textbf{Avg. ST:} the average solution time required by each dispatching method for within a dispatching wave.
7) \textbf{Avg. WL:} the average workload-imbalance ratio across couriers after dispatching. 
At each wave, the workload of courier $k$ is measured by 
$w_k = |O_k^{old}|+\sum_{i\in O^{new}} y_i^{k}$. 
We define
$
\mathrm{WL}=\frac{\mathrm{std}_{k\in\mathcal{K}}(w_k)}{\mathrm{mean}_{k\in\mathcal{K}}(w_k)},
$
and report its average over all waves. 
This metric reflects how evenly the active workload is distributed among couriers, while smaller values indicate a more balanced dispatch.

\begin{table*}[ht]
\centering
\footnotesize
\caption{Online rolling-horizon simulation results for Shanghai, Hangzhou, Chongqing, Jilin, and Yantai}
\label{online_combined}
\begin{tabular}{lccccccc}
\toprule
\multicolumn{8}{c}{\textbf{Shanghai ($N=20, K=9$)}} \\
\midrule
Methods & Avg. Obj & Avg. TWVR & Avg. TWP & Max. TWP  & Avg. TT & Avg. ST & Avg. WL \\
\midrule
DMCA            & 7.38 & 0.84 & 2.14 & 0.86 & 7.49 & 8.79   & 0.79 \\
Greedy          & 6.58 & 0.62 & 1.86 & 0.74 & 6.74 & \textbf{3.56}   & 0.44 \\
PP-Greedy       & 6.32 & 0.60 & 1.74 & 0.68 & 6.54 & 3.92   & 0.41 \\
TS-Dispatching  & 6.11 & 0.25 & 1.58 & \textbf{0.44} & 6.47 & 108.13 & 0.28 \\
PP-Greedy-LS    & \textbf{5.82} & \textbf{0.22} & \textbf{1.34} & 0.46 & \textbf{6.40} & 20.91  & \textbf{0.21} \\
\midrule
\multicolumn{8}{c}{\textbf{Hangzhou ($N=22, K=10$)}} \\
\midrule
DMCA            & 12.58 & 0.42 & 3.42 & 1.14 & 13.09 & 6.32   & 0.58 \\
Greedy          & 11.82 & 0.38 & 3.12 & 1.02 & 12.43 & \textbf{3.14}   & 0.51 \\
PP-Greedy       & 10.46 & 0.33 & 2.74 & 0.88 & 11.03 & 3.28   & 0.44 \\
TS-Dispatching  & 9.84  & 0.28 & 2.41 & 0.64 & 10.61 & 62.45  & 0.38 \\
PP-Greedy-LS    & \textbf{9.12} & \textbf{0.19} & \textbf{2.16} & \textbf{0.58} & \textbf{9.94}  & 28.12  & \textbf{0.31} \\
\midrule
\multicolumn{8}{c}{\textbf{Chongqing ($N=16, K=10$)}} \\
\midrule
DMCA            & 31.43 & 0.44 & 8.42 & 1.68 & 32.87 & 10.85  & 0.52 \\
Greedy          & 30.15 & 0.41 & 7.82 & 1.42 & 31.90 & \textbf{2.24}   & 0.46 \\
PP-Greedy       & 28.44 & 0.36 & 7.14 & 1.28 & 30.43 & 5.12   & 0.38 \\
TS-Dispatching  & 27.22 & 0.28 & 6.84 & 1.04 & 29.11 & 78.41  & 0.31 \\
PP-Greedy-LS    & \textbf{26.39} & \textbf{0.20} & \textbf{6.24} & \textbf{0.92} & \textbf{28.79} & 18.56  & \textbf{0.25} \\
\midrule
\multicolumn{8}{c}{\textbf{Jilin ($N=19, K=11$)}} \\
\midrule
DMCA            & 23.41 & 0.42 & 6.24 & 1.32 & 24.53 & 12.14  & 0.78 \\
Greedy          & 22.63 & 0.38 & 5.84 & 1.18 & 23.99 & \textbf{2.22}   & 0.72 \\
PP-Greedy       & 21.84 & 0.31 & 5.34 & 1.04 & 23.57 & 5.84   & 0.62 \\
TS-Dispatching  & 21.02 & 0.22 & 4.96 & \textbf{0.84} & 22.94 & 84.62  & 0.54 \\
PP-Greedy-LS    & \textbf{20.15} & \textbf{0.12} & \textbf{4.62} & 0.86 & \textbf{22.19} & 23.42  & \textbf{0.38} \\
\midrule
\multicolumn{8}{c}{\textbf{Yantai ($N=22, K=10$)}} \\
\midrule
DMCA            & 38.45 & 0.52 & 10.32 & 1.84 & 40.19 & 13.12  & 0.72 \\
Greedy          & 35.12 & 0.48 & 9.44  & 1.58 & 36.69 & \textbf{3.56}   & 0.64 \\
PP-Greedy       & 32.68 & 0.42 & 8.64  & 1.42 & 34.34 & 4.12   & 0.54 \\
TS-Dispatching  & 30.24 & 0.31 & 7.54  & 1.18 & 32.43 & 115.42 & 0.42 \\
PP-Greedy-LS    & \textbf{28.56} & \textbf{0.22} & \textbf{6.84} & \textbf{0.98} & \textbf{31.03} & 28.36  & \textbf{0.35} \\
\bottomrule
\end{tabular}
\end{table*}

Table \ref{online_combined} provides a comprehensive evaluation of PP-Greedy-LS compared with other baselines across five cities. 
The results demonstrate that PP-Greedy-LS consistently outperforms other baselines across all of the instances from five cities. 
Within an acceptable increase in runtime of ($\approx30$ seconds), PP-Greedy-LS significantly reduces the average objective compared to PP-Greedy, such as in the Yantai instance where the objective improves from 32.68 to 28.56. This superior performance is achieved through shorter Avg. TT and lower Avg. TWP, with the latter dropping to 2.16 in Hangzhou and 4.62 in Jilin, which underscores the practicality and necessity of the local search component in dispatching optimization.
Notably, Avg. WL results indicate that dispatching decisions under PP-Greedy-LS are more equitably distributed among couriers, with imbalance ratios falling to 0.21 in Shanghai and 0.25 in Chongqing. This distribution prevents over-dispatching to individual couriers and further reduces the Max. TWP from 1.14 for DMCA in Hangzhou to only 0.58 for PP-Greedy-LS, thereby providing robust real-time decision support for logistics platforms.

\subsubsection{The impact of supply-to-demand ratio} \label{exp_4}
In real-world pickup ecosystems, dispatching outcomes are fundamentally governed by the dynamic interplay between the temporal fluctuations of the number of new orders $|O^{new}|$ (demand) and the number of candidate couriers $K$ (supply). To evaluate the robustness and scalability of the proposed PP-Greedy-LS under varying operational pressures, we next conduct a sensitivity analysis based on the Supply-to-Demand Ratio (SDR), defined as the ratio of candidate couriers to new orders. This analysis is motivated by the need to ensure that the proposed algorithm maintains a competitive advantage not only during resource-abundant off-peak periods but also during highly congested peak intervals, where the tighter coupling of constraints typically leads to a degradation in service quality for less sophisticated heuristics. 
Specifically, to isolate the impact of the supply-demand ratio, we select a representative dispatching instance from the Shanghai dataset with $|O^{new}|=20$ and vary $K$ to derive different SDR. The corresponding values for Obj, TWVR, TWP, Max. TWP, TT, Avg. ST, and WL are reported in Table \ref{SDR}. These results characterize the system's performance across a broad spectrum of operational conditions, spanning from severe peak-demand pressure ($SDR=0.2$) to significant supply abundance ($SDR=2.0$).

\begin{table*}[htbp]
\centering
\footnotesize
\caption{Impact of SDR on system performance for Shanghai instance ($N=20, \alpha=0.7, \phi=1.0$)}
\label{SDR}
\begin{tabular}{cccccccccc}
\toprule
$|O^{new}|$ & $K$ & SDR & Obj & TWVR & TWP & Max. TWP & TT & Avg. ST  & WL \\
\midrule
20 & 4  & 0.2 & 16.28 & 0.82 & 9.81 & 3.42 & 9.24 & 14.07 & 0.65 \\
20 & 8  & 0.4 & 13.24 & 0.65 & 7.33 & 2.56 & 8.44 & 16.23 & 0.52 \\
20 & 12 & 0.6 & 11.35 & 0.52 & 5.88 & 2.05 & 7.82 & 17.84 & 0.44 \\
20 & 16 & 0.8 & 9.46  & 0.42 & 4.32 & 1.51 & 7.34 & 19.46 & 0.38 \\
20 & 20 & 1.0 & 9.08  & 0.38 & 4.10 & 1.44 & 7.12 & 20.45 & 0.35 \\
20 & 24 & 1.2 & 7.54  & 0.25 & 2.82 & 0.98 & 6.74 & 22.08 & 0.28 \\
20 & 28 & 1.4 & 6.91  & 0.18 & 2.35 & 0.82 & 6.52 & 23.91 & 0.25 \\
20 & 32 & 1.6 & 6.55  & 0.15 & 2.11 & 0.74 & 6.34 & 25.18 & 0.22 \\
20 & 36 & 1.8 & 6.32  & 0.12 & 1.97 & 0.68 & 6.22 & 27.06 & 0.21 \\
20 & 40 & 2.0 & 6.07  & 0.10 & 1.79 & 0.62 & 6.12 & 29.02 & 0.19 \\
\bottomrule
\end{tabular}
\end{table*}

\begin{figure}[htb!]
	\centering
	\includegraphics[scale=0.45]{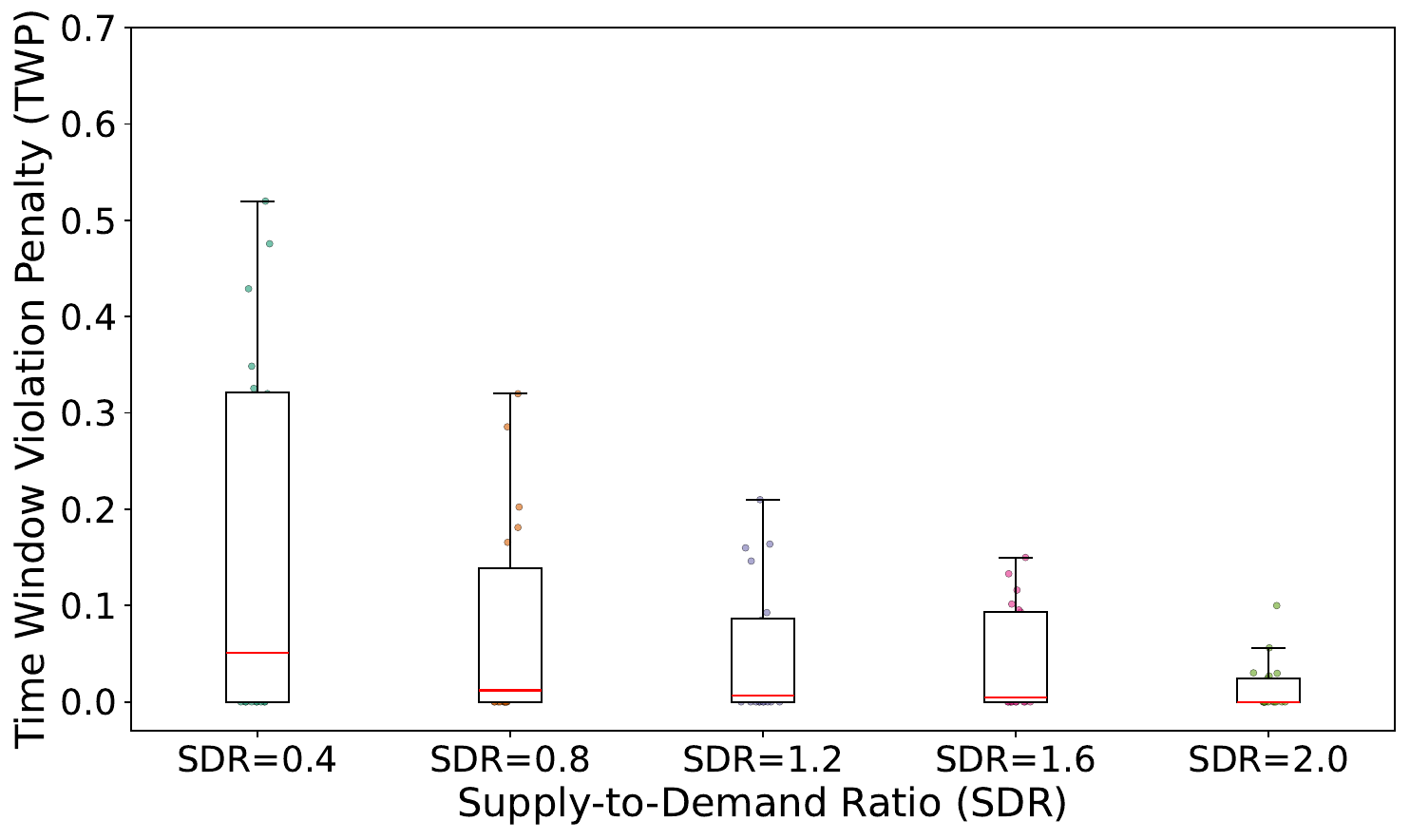}
	\caption{The impact of SDR on time window violation penalties}
	\label{SDR_box}
\end{figure}

Table~\ref{SDR} shows that when the system is supply-constrained (low SDR, e.g., SDR= 0.2 and 0.4), infeasibilities and TWP dominates the objective, whereas increasing SDR progressively shifts the system into a regime where TT becomes the primary driver of performance. 
Importantly, the effect of SDR exhibits diminishing returns: increasing SDR from a low level yields substantial gains, whereas further increases beyond a moderate level lead to only incremental improvements.
For instance, increasing SDR from $0.2$ to $1.0$ reduces the objective from $16.28$ to $9.08$ and lowers $TWVR$ from $0.82$ to $0.38$, indicating that additional supply primarily mitigates TWP and stabilizes service quality. Once the system has sufficient coverage, the marginal benefit diminishes. Increasing SDR from $1.4$ to $2.0$ improves the objective only from $6.91$ to $6.07$, suggesting that extra candidates mainly refine routing decisions rather than recovering feasibility, where Figure \ref{SDR_box} also shows consistent results.

From the perspective of logistics platforms, this supports a two-regime operating policy. 
During peak hours, prioritize supply-side interventions such as temporary courier activation, short-term incentives, or enlarging the candidate set via geo-fencing to avoid violation-driven degradation. 
During off-peak periods, allocate effort to routing and coordination measures such as tighter spatial matching or zoning, because expanding the candidate pool yields only limited incremental gains.

\subsubsection{Ablation study} \label{exp_5}
The DR-LaCPNet model proposed in this paper includes a DRGAN encoder and a LaCP decoder, where LaCP integrates a look-ahead travel-time encoding (LA) and a courier-personalization module (CP). 
In order to verify the rationality of adding these components, we made partial changes to the DR-LaCPNet model framework and conducted ablation studies. 
The ablation study evaluates the performances of the LaCPNet (DR-LaCPNet without the DRGAN Encoder), DR-Net (DR-LaCPNet without the LaCP decoder), DR-CPNet (DR-LaCPNet without the Look-Ahead mechanism), DR-LaNet (DR-LaCPNet without the CP mechanism).
We assess the performance degradation on the instances in Section~\ref{exp_1} for routing-only optimization and on the instances in Section~\ref{exp_2} for the integrated dispatching and routing optimization, respectively.
The columns ``Obj Imp.'' report the relative improvement of other networks compared with LaCPNet.

\begin{table*}[htbp]
\centering
\footnotesize
\caption{Performance comparison under different scenarios}
\resizebox{\textwidth}{!}{%
\begin{tabular}{cc cc ccc cc cc ccc}
\toprule
\multicolumn{2}{c}{Scenario}
& \multicolumn{2}{c}{LaCPNet}
& \multicolumn{3}{c}{DR-Net}
& \multicolumn{2}{c}{DR-CPNet}
& \multicolumn{2}{c}{DR-LaNet}
& \multicolumn{3}{c}{DR-LaCPNet} \\
\cmidrule(lr){1-2}
\cmidrule(lr){3-4}
\cmidrule(lr){5-7}
\cmidrule(lr){8-9}
\cmidrule(lr){10-11}
\cmidrule(lr){12-14}
$N$ & $\delta$
& Obj & 
& Obj &  & Obj Imp.
& Obj & Obj Imp.
& Obj & Obj Imp.
& Obj &  & Obj Imp. \\
\midrule

\multirow{3}{*}{$[1,4]$}
& 1 & 0.53 &  & 0.53 &  & 0.00  & 0.53 & 0.00  & 0.53 & 0.00  & \textbf{0.50} &  & 5.66 \\
& 3 & 0.34 &  & 0.37 &  & -8.82 & 0.34 & 0.00  & 0.34 & 0.00  & \textbf{0.31} &  & 8.82 \\
& 5 & 0.20 &  & 0.20 &  & 0.00  & 0.20 & 0.00  & 0.20 & 0.00  & \textbf{0.19} &  & 5.00 \\
\addlinespace

\multirow{3}{*}{$[5,8]$}
& 1 & 1.83 &  & 1.40 &  & 23.50 & 1.20 & 34.43 & 1.20 & 34.43 & \textbf{1.17} &  & 36.07 \\
& 3 & 1.99 &  & 1.70 &  & 14.57 & 1.64 & 17.59 & 1.61 & 19.10 & \textbf{1.36} &  & 31.66 \\
& 5 & 1.16 &  & 1.23 &  & -6.03 & 1.16 & 0.00  & 1.23 & -6.03 & \textbf{1.11} &  & 4.31 \\
\addlinespace

\multirow{3}{*}{$[9,12]$}
& 1 & 1.63 &  & 1.58 &  & 3.07  & 1.74 & -6.75 & 1.58 & 3.07  & \textbf{1.52} &  & 6.75 \\
& 3 & 1.72 &  & 1.80 &  & -4.65 & 1.75 & -1.74 & 1.48 & 13.95 & \textbf{1.41} &  & 18.02 \\
& 5 & 2.07 &  & 2.15 &  & -3.86 & 2.07 & 0.00  & 1.90 & 8.21  & \textbf{1.55} &  & 25.12 \\
\addlinespace

\multirow{3}{*}{$[13,16]$}
& 1 & 5.43 &  & 3.60 &  & 33.70 & 4.08 & 24.86 & 4.78 & 11.97 & \textbf{3.15} &  & 41.99 \\
& 3 & 4.06 &  & 3.19 &  & 21.43 & 3.07 & 24.38 & 2.95 & 27.34 & \textbf{2.72} &  & 33.00 \\
& 5 & 2.80 &  & 2.45 &  & 12.50 & 2.52 & 10.00 & 2.74 & 2.14  & \textbf{2.12} &  & 24.29 \\

\bottomrule
\end{tabular}%
}
\label{tab_ablation1}
\end{table*}

Table~\ref{tab_ablation1} shows that DR-LaCPNet consistently delivers the best objective across all scenarios, and the benefit of the full design grows with instance scale. 
For small instances ($N\in[1,4]$), all variants are close, indicating limited room for architectural enhancements; however, as $N$ increases, removing either the DRGAN encoder (LaCPNet) or the LaCP decoder (DR-Net) leads to clear degradation, while the full model maintains a stable advantage (e.g., $N\in[13,16]$, the improvement reaches up to 41.99 \% at $\delta=1$). 
Moreover, ablations inside LaCP confirm that both the look-ahead mechanism and courier-personalized mechanism contribute to the final performance, with their impact becoming more pronounced on harder, larger-scale settings, suggesting that anticipating downstream travel-time effects and capturing courier heterogeneity are both crucial to sustaining solution quality when the routing problem becomes complex.

\subsubsection{The impact of objective weights} \label{exp_6}
To better characterize how objective weights design translates into operational behavior, we conduct a sensitivity analysis by varying $(\alpha,\phi)$. 
In practice, these weights are policy levers that shift the system from a service-oriented regime to an efficiency-oriented regime. 
This analysis is crucial because real-world pickup operations often experience regime shifts across time and regions, where managers must deliberately re-tune priorities to meet different KPIs. 
Moreover, service quality and efficiency can trade off nonlinearly under tight time windows, so a weight sweep provides a transparent way to quantify the resulting managerial trade-offs and assess how robust each method remains across regimes. 
Therefore, we vary $\alpha$ on the Shanghai instance with $N=20$ and $K=9$ over a set of representative configurations spanning from service-oriented to efficiency-oriented settings, and report the resulting objective values and operational metrics for each configuration.

\begin{table*}[htbp]
\centering
\footnotesize
\caption{Sensitivity analysis for objective weights}
\label{managerial_insights_table}
\begin{tabular}{lccccccc}
\toprule
Method & Avg. obj & Avg. TWVR & Avg. TWP & Max. TWP & Avg. TT & Avg. ST & Avg. WL \\
\midrule
\multicolumn{8}{c}{\textbf{Scenario A: Service-Oriented ($\alpha=0.3$, $\phi=1.0$)}} \\
\midrule
DMCA            & 4.54 & 0.28 & 1.48 & 0.44 & 10.21 & 8.74  & 0.76 \\
Greedy          & 4.12 & 0.24 & 1.27 & 0.41 & 9.49  & 3.52  & 0.81 \\
PP-Greedy       & 3.84 & 0.21 & 1.12 & 0.36 & 9.07  & 3.88  & 0.73 \\
TS-Dispatching  & 3.66 & 0.17 & 0.92 & \textbf{0.27} & 9.14  & 108.24 & 0.29 \\
PP-Greedy-LS    & \textbf{3.34} & \textbf{0.13} & \textbf{0.83} & 0.29 & \textbf{8.37} & 20.84 & \textbf{0.22} \\
\midrule
\multicolumn{8}{c}{\textbf{Scenario B: Balanced ($\alpha=0.7$, $\phi=1.0$)}} \\
\midrule
DMCA            & 7.38 & 0.84 & 2.14 & 0.86 & 7.49 & 8.79   & 0.79 \\
Greedy          & 6.58 & 0.62 & 1.86 & 0.74 & 6.74 & 3.56   & 0.44 \\
PP-Greedy       & 6.32 & 0.60 & 1.74 & 0.68 & 6.54 & 3.92   & 0.41 \\
TS-Dispatching  & 6.11 & 0.25 & 1.58 & \textbf{0.44} & 6.47 & 108.13 & 0.28 \\
PP-Greedy-LS    & \textbf{5.82} & \textbf{0.22} & \textbf{1.34} & 0.46 & \textbf{6.40} & 20.91  & \textbf{0.21} \\
\midrule
\multicolumn{8}{c}{\textbf{Scenario C: Efficiency-Oriented ($\alpha=1.1$, $\phi=1.0$)}} \\
\midrule
DMCA            & 9.87 & 0.94 & 5.16 & 1.82 & 4.28 & 8.81  & 0.77 \\
Greedy          & 8.92 & 0.90 & 4.67 & 1.64 & 3.86 & 3.54  & 0.81 \\
PP-Greedy       & 8.41 & 0.84 & 4.38 & 1.52 & 3.66 & 3.94  & 0.74 \\
TS-Dispatching  & 7.94 & 0.80 & 4.14 & 1.14 & 3.45 & 107.92 & 0.73 \\
PP-Greedy-LS    & \textbf{7.46} & \textbf{0.78} & \textbf{3.92} & \textbf{1.10} & \textbf{3.22} & 21.04 & \textbf{0.74} \\
\bottomrule
\end{tabular}
\end{table*}

Table~\ref{managerial_insights_table} shows a clear shift from service-oriented to efficiency-oriented operations as the objective weights change. 
For PP-Greedy-LS, moving from Scenario A to Scenario C compresses travel time substantially (Avg. TT drops from 8.37 to 3.22) but increases time-window penalties (Avg. TWP rises from 0.83 to 3.92 and Max. TWP from 0.29 to 1.10).
This suggests a practical lever for managers: use more service-oriented weights in peak or high-penalty contexts to cap violation risk, and switch toward efficiency-oriented weights in off-peak or capacity-abundant periods to prioritize throughput. 
Across all regimes, PP-Greedy-LS remains the most robust, achieving the best objective in Scenarios A–C while keeping violations low, whereas TS-Dispatching attains comparable service metrics but at much higher latency (Avg. ST $\approx 108s$ vs. $\approx21s$), making it less suitable for frequent real-time re-optimization.

\section{Conclusion} \label{conclu}
In this work, we study the integrated order dispatching and routing problem arising in the last-mile pickup. 
To address this challenging problem, we propose an integrated optimization framework which couples a learned routing oracle with real-time dispatching heuristics. We first formulate the dispatch problem and routing problem as two MILP models, incorporating time-dependent travel time. For the routing subproblem, we initially model it as a sequential decision Markov Decision Process. Subsequently, we propose a DR-LaCPNet framework comprising a Dynamic-Residual Graph Attention Network encoder, and a Look-Ahead Courier-Personalized decoder. While the Look-Ahead mechanism effectively handles the uncertainty introduced by time-dependent travel times, the courier-personalized mechanism accounts for courier heterogeneity. 
We then introduce Position-pool Greedy Dispatch with Local Search Refinement for order dispatching, where we innovatively integrate the trained DRL routing oracle into this heuristic to rapidly screen dispatchable couriers.

Subsequently, we conduct extensive experiments based on the  LADE dataset of Cainiao Logistics.
We first propose the dispatch instance simulator leveraging the spatial-temporal characteristics of the LADE dataset to construct effective training and testing instances, and further validate the model's convergence. 
Following this, we evaluate the performance of the DRL routing oracle. 
Extensive experiments demonstrate the effectiveness of the DRL routing oracle compared against other heuristics and deep learning models. 
Subsequently, the proposed integrated optimization framework demonstrates outstanding performance in both offline evaluation and online rolling-horizon simulation, yielding satisfactory solutions within short decision times. 
We further examine the impact of the supply-to-demand ratio on the solution and reveal the role of each component through ablation studies. 
Finally, by training multiple DRL models with different objective weights, we compare the consequent results and provide managerial insights for logistics platforms.

For the future work, one possible direction is to address the issues of overly sparse training rewards and the inability of graph-based training to generalize across varying numbers of nodes. 
This would enable more effective strategies for integrating order dispatch and routing, such as adopting hierarchical deep reinforcement learning. 
Another possible direction is that while the current model only solves the last-mile pickup problem, the proposed look-ahead mechanism and courier personalization mechanism demonstrate broad applicability and can be extended to other similar routing problems. 
Finally, utilizing heterogeneous graphs to distinguish order nodes from courier nodes for enhanced feature extraction represents a promising future direction.

\section*{Acknowledgments}
This research was supported by the National Natural Science Foundation Council of China under Projects 72361137004, 92567301 and 92167206.

\begin{singlespace}
\bibliographystyle{apalike}
\bibliography{reference}
\end{singlespace}

\newpage
\appendix
\appendixpage
\section{Baseline heuristics}\label{append_1}
The baseline heuristics used for comparison in Sections \ref{exp_2}, \ref{exp_3}, and \ref{exp_6} are summarized as follows.
\begin{algorithm}[ht]
	\renewcommand{\algorithmicrequire}{\textbf{Input:}}
	\renewcommand{\algorithmicensure}{\textbf{Output:}}
	\caption{Greedy dispatch heuristic (Greedy)}
	\begin{algorithmic}[1]
	\STATE Input: new orders $O^{new}$, candidate couriers $\mathcal{K}$, trained routing policy $\pi^R$, candidate size $Q$
	\STATE Compute $J_{\text{old}}^{k}$ for all $k\in\mathcal{K}$ using $\pi^R$; set $J_{\text{tot}} \leftarrow \sum_{k\in\mathcal{K}} J_{\text{old}}^{k}$
	\STATE Sort orders in $O^{new}$ in non-descending order by urgency $b_i$ to obtain $\hat{O}^{new}$
	\FOR{each order $i \in \hat{O}^{new}$}
	\STATE Select the top-$Q$ candidate couriers in $\mathcal{K}$ with the smallest Euclidean distance to order $i$ (based on current courier locations), denoted by $\mathcal{K}_Q(i)$
	\STATE $k^\star \leftarrow \arg\min_{k\in\mathcal{K}_Q(i)} \Delta_{ik}$, where $\Delta_{ik}=\text{GreedyRolloutCost}\!\left(\pi^{R},\, O^{old}_{k}\cup\{i\}\right)-J_{\text{old}}^{k}$
	\IF{$k^\star$ exists}
	\STATE Set $y_i^{k^\star} \leftarrow 1$ and $y_i^{k} \leftarrow 0$ for all $k\neq k^\star$
	\STATE $J_{\text{old}}^{k^\star} \leftarrow J_{\text{old}}^{k^\star} + \Delta_{ik^\star}$, $J_{\text{tot}} \leftarrow J_{\text{tot}} + \Delta_{ik^\star}$
	\ELSE
	\STATE Continue
	\ENDIF
	\ENDFOR
	\STATE Output: $\{y_i^k\}_{i\in O^{new},k\in\mathcal{K}}$ and $J_{\text{tot}}$
	\end{algorithmic}
    \label{alg_append1}
\end{algorithm}

We first use the greedy dispatch heuristic (Greedy) as the baseline. As shown in Algorithm \ref{alg_append1}, Greedy first evaluates each courier's current routing cost $J_{\text{old}}^k$ under $\pi^R$ and aggregates them to obtain the initial system cost $J_{\text{tot}}=\sum_{k \in \mathcal{K}} J_{\text{old}}^k$. The orders in $O^{new}$ are then sorted in a non-descending order of urgency $b_i$, yielding $\hat{O}^{new}$, so that more urgent orders are processed first. For each order $i \in \hat{O}^{new}$, the heuristic restricts the dispatch candidates to a local neighborhood by selecting $\mathcal{K}_Q(i)$, the top-$Q$ couriers with the smallest Euclidean distance to $i$ (based on their current locations). The heuristic then chooses the courier $k^{\star}$ that minimizes the incremental cost $\Delta_{ik}$, defined as the difference between the rollout routing cost after tentatively inserting $i$ into courier $k$'s workload and the current cost $J_{\text{old}}^k$, i.e.,
$\Delta_{ik}=\operatorname{GreedyRolloutCost}\!\left(\pi^R,\, O^{old}_k\cup\{i\}\right)-J_{\text{old}}^k$.
If a feasible $k^{\star}$ exists, the corresponding dispatch variable is set to $y_i^{k^\star}=1$, and both $J_{\text{old}}^{k^\star}$ and $J_{\text{tot}}$ are updated by adding $\Delta_{ik^\star}$. Orders that cannot be feasibly assigned are skipped. The heuristic outputs the dispatch decisions $\left\{y_i^k\right\}$ for all $i \in O^{new}$ and the resulting total routing cost $J_{\text{tot}}$.

\begin{algorithm}[ht]
	\renewcommand{\algorithmicrequire}{\textbf{Input:}}
	\renewcommand{\algorithmicensure}{\textbf{Output:}}
	\caption{Position-pool greedy dispatch heuristic (PP-Greedy)}
	\begin{algorithmic}[1]
	\STATE Input: new orders $O^{new}$, candidate couriers $\mathcal{K}$, trained routing policy $\pi^R$, candidate size $Q$
	\STATE Compute $J_{\text{old}}^{k}$ for all $k\in\mathcal{K}$ using $\pi^R$; set $J_{\text{tot}} \leftarrow \sum_{k\in\mathcal{K}} J_{\text{old}}^{k}$
	\STATE Initialize $O^{new}_{k}\leftarrow \emptyset$ for all $k\in\mathcal{K}$
	\STATE Initialize the position pool $S_k \leftarrow \{l_k\}\cup \{l_j: j\in O^{old}_{k}\}$ for all $k\in\mathcal{K}$
	\STATE Sort orders in $O^{new}$ in non-descending order by urgency $b_i$ to obtain $\hat{O}^{new}$
	\FOR{each order $i \in \hat{O}^{new}$}
		\STATE Compute $d_{ik}\leftarrow \min_{p\in S_k}\|l_i-p\|_2$ for all $k\in\mathcal{K}$
		\STATE Select the top-$Q$ candidate couriers with the smallest $d_{ik}$, denoted by $\mathcal{K}^{\text{pool}}_Q(i)$
		\STATE $k^\star \leftarrow \arg\min_{k\in\mathcal{K}^{\text{pool}}_Q(i)} \Delta_{ik}$, where
		$\Delta_{ik}=\text{GreedyRolloutCost}\!\left(\pi^{R},\, O^{old}_{k}\cup O^{new}_{k}\cup\{i\}\right)-J_{\text{old}}^{k}$
		\IF{$k^\star$ exists}
			\STATE Set $y_i^{k^\star} \leftarrow 1$ and $y_i^{k} \leftarrow 0$ for all $k\neq k^\star$
			\STATE Update $O^{new}_{k^\star}\leftarrow O^{new}_{k^\star}\cup\{i\}$ and $S_{k^\star}\leftarrow S_{k^\star}\cup\{l_i\}$
			\STATE Update $J_{\text{old}}^{k^\star} \leftarrow J_{\text{old}}^{k^\star} + \Delta_{ik^\star}$, $J_{\text{tot}} \leftarrow J_{\text{tot}} + \Delta_{ik^\star}$
		\ELSE
			\STATE Continue
		\ENDIF
	\ENDFOR
	\STATE Output: $\{y_i^k\}_{i\in O^{new},k\in\mathcal{K}}$ and $J_{\text{tot}}$
	\end{algorithmic}
    \label{alg_append2}
\end{algorithm}

Another baseline model we have employed is the position-pool greedy dispatch heuristic (PP-Greedy), which excludes the local search refinement. As shown in Algorithm \ref{alg_append2}, PP-Greedy follows the same routing-aware greedy rollout framework but replaces the candidate-screening rule by a position-pool mechanism. Specifically, for each courier $k$, we maintain a position pool $S_k$ that contains the courier's current position $l_k$ and the positions of its already assigned stops (i.e., the pickup locations of orders in $O^{old}_k$), and we further augment $S_k$ on-the-fly as new orders are dispatched to $k$. For each incoming order $i$, we compute its distance to courier $k$ as the minimum Euclidean distance to the pool,
$d_{ik}=\min_{p\in S_k}\|l_i-p\|_2$,
and select $\mathcal{K}_Q^{\text{pool}}(i)$, the top-$Q$ couriers with the smallest $d_{ik}$, as a local candidate set. We then evaluate the incremental cost for each candidate $k\in\mathcal{K}_Q^{\text{pool}}(i)$ using the routing policy $\pi^R$ via a greedy rollout, i.e.,
$\Delta_{ik}=\operatorname{GreedyRolloutCost}\!\left(\pi^R,\, O^{old}_{k}\cup O^{new}_{k}\cup\{i\}\right)-J_{\text{old}}^{k}$,
and dispatch $i$ to the courier $k^\star$ that yields the minimum $\Delta_{ik}$. Compared with screening candidates solely by the courier's current position, the position pool provides a more robust spatial proxy of a courier's near-term service footprint, since it accounts for proximity to any already-committed stop. After assigning $i$ to $k^\star$, PP-Greedy updates $O^{new}_{k^\star}$, augments $S_{k^\star}$ with $l_i$, and updates both $J_{\text{old}}^{k^\star}$ and the system cost $J_{\text{tot}}$ accordingly.

\end{document}